\documentclass[final,5p,times,twocolumn]{elsarticle}

\usepackage[ruled]{algorithm2e}
\usepackage{tabularx}
\usepackage{amssymb}
\usepackage{amsmath}
\usepackage{stfloats}
\usepackage{ulem}
\usepackage{graphicx}
\usepackage{subfigure}
\usepackage{booktabs}
\usepackage{bbding}
\usepackage{multirow}
\usepackage{makecell}
\usepackage{array}
\AtBeginEnvironment{algorithm}

\journal{arxiv}

\begin{document}

\begin{frontmatter}

\title{Edge-assisted U-Shaped Split Federated Learning with Privacy-preserving for Internet of Things}

\author[1]{Hengliang Tang}
\ead{tanghengliangbwu@163.com}
\author[1]{Zihang Zhao}
\ead{zhaozihang1998@foxmail.com}
\author[2]{Detian Liu}
\ead{liudt2021@126.com}
\author[1]{Yang Cao}
\ead{caoyangcwz@126.com}
\author[1]{Shiqiang Zhang}
\ead{sqzhangbjut@163.com}
\author[1]{Siqing You\corref{cor1}}
\ead{yousiqing0203@126.com}

\cortext[cor1]{Corresponding author}
\address[1]{School of Information, Beijing Wuzi University, Beijing, 101149, China}
\address[2]{Faculty of Information Technology, Beijing University of Technology, Beijing, 100124, China}

\begin{abstract}
In the realm of the Internet of Things (IoT), deploying deep learning models to process data generated or collected by IoT devices is a critical challenge. However, direct data transmission can cause network congestion and inefficient execution, given that IoT devices typically lack computation and communication capabilities. Centralized data processing in data centers is also no longer feasible due to concerns over data privacy and security. To address these challenges, we present an innovative Edge-assisted U-Shaped Split Federated Learning (EUSFL) framework, which harnesses the high-performance capabilities of edge servers to assist IoT devices in model training and optimization process. In this framework, we leverage Federated Learning (FL) to enable data holders to collaboratively train models without sharing their data, thereby enhancing data privacy protection by transmitting only model parameters. Additionally, inspired by Split Learning (SL), we split the neural network into three parts using U-shaped splitting for local training on IoT devices. By exploiting the greater computation capability of edge servers, our framework effectively reduces overall training time and allows IoT devices with varying capabilities to perform training tasks efficiently. Furthermore, we proposed a novel noise mechanism called LabelDP to ensure that data features and labels can securely resist reconstruction attacks, eliminating the risk of privacy leakage. Our theoretical analysis and experimental results demonstrate that EUSFL can be integrated with various aggregation algorithms, maintaining good performance across different computing capabilities of IoT devices, and significantly reducing training time and local computation overhead. 
\end{abstract}




\end{frontmatter}

\section{Introduction}

The fast-paced growth and widespread adoption of IoT have resulted in the increased usage of IoT devices in various scenarios\cite{empoweringAI}. For instance, sensor devices are commonly deployed in intelligent detection systems, and IoT devices, together with technologies like deep learning and artificial intelligence, are used in joint applications. Meanwhile, IoT devices are widely distributed, providing an extensive source of data support for AI. According to Ericsson's prediction, the total amount of internet data will reach 40ZB in 2024, with IoT devices contributing 45\% of the total data generated. As the amount of data generated by IoT continues to rise, the conventional approach of centralizing data is no longer feasible \cite{distributedlearning}. Transmitting data can lead to difficulties such as network congestion and excessive transmission. Additionally, unauthorized data manipulation can result in legal issues due to the constraints of data privacy protection regulations like GDPR\cite{gdpr}. Therefore, it is crucial to investigate new technical solutions that incorporate privacy protection features suitable for the IoT environment.

With the rising concern on privacy, Federated Learning (FL), which is a distributed machine learning architecture with data privacy protection feature, is considered as a promising solution for IoT scenarios to extract the potential of data produced by IoT devices without the risk of privacy leakage\cite{kairouz2021advances}. Unlike traditional distributed machine learning, FL only requires the participants to upload the trained local model update to perform aggregation to obtain a global model that can be used for sharing without any transfer of raw data, eliminating the risk of privacy leakage when data holders distribute or transmit data \cite{FL-source}. Thus, within the IoT environments, the raw data is utilized by IoT devices for local model training, after which only the local model parameters are uploaded for global model aggregation.

As AI-related research progresses, the scale of machine learning and deep learning models continues to expand. Through the analysis of current computational requirements for SOTA models, it has been found that the model parameters can reach up to 732M, 727M, and 23.8B in different fields such as speech recognition, image classification, and natural language processing \cite{bigmodel}. While deploying these functions in IoT environment, it is clear that models on IoT devices requires higher computational capabilities. Despite the growing performance of IoT devices, they still face challenges when dealing with such immense and rapidly increasing computational complexities and may have an impact on other tasks on the device \cite{IoTdlproblem}. More new approaches need to be explored to enable IoT devices, which are relatively weak in computational capability, to effectively deploy models to perform training.

Split learning (SL) is a powerful technique that enables the distributed training of neural network models by allowing different layers to be allocated to different entities for training and inference purposes\cite{abuadbba2020can}. Common approach involves layer-based model splitting between client and server, whereby model training is divided into two parts. The first part is performed on the client, with the second part executed on the server. SL can effectively alleviate the computational burden on IoT devices, thereby facilitating model training on resource-constrained devices, and also providing decent privacy guarantee.

Given the advantages of both FL and SL in deep learning model training, it is a logical step to combine these techniques to improve the efficiency of IoT devices in the training process. By integrating FL and SL, it is possible to leverage the data privacy-preserving properties of FL with the model privacy properties of SL, thereby ensuring privacy for both data and models \cite{fromFLtoSL}. The main approach is to split an intact neural network model into multiple parts based on layers, to have the computationally demanding parts of the split model done on the server side with more computational power, and for IoT devices to train only the low-computational parts of the model. 

However, introducing these methods into practical scenarios also brings new threats. Therefore, new security analysis and protective measures need to be considered. We need to ensure that our framework is secure against attacks. Therefore, we proposed LabelDP, a new noise mechanism, further enhance the privacy protection capabilities of our framework. 

According to the above description, to address the problems of data privacy and insufficient computing capability of devices in the IoT environment, this article proposes an edge-assisted U-shaped split federated learning (EUSFL) framework, which exploit edge servers in a three-tier architecture of "Client-Edge-Central" to assist IoT devices to complete the training process of local models, improve the overall computational efficiency, reduce the computational overhead of devices, and maintain data privacy. In EUSFL framework, IoT devices can use their local data for local model training, and U-shaped split learning method make it applicable to the IoT environment
to make full use of the computing resources of the edge servers and devices to complete the training and inference process more efficiently. The U-shaped splitting approach involves splitting the neural network model into three parts. The middle part, which requires more computational power, is supported by the edge server, while the first and last parts are performed on the IoT devices. This approach entails the transmission of only the parameters obtained from each part of the training, ensuring that each part of the model remains inaccessible to the others and thereby improving model privacy. Hence, within the EUSFL framework, it is feasible to maintain the privacy of raw data held by the data holder, while also safeguarding the label information in model training, resulting in a privacy-preserving design with excellent computational efficiency. Additionally, this approach offers superior scalability compared to both the vanilla SL method and traditional FL methods.

In summary, the main contributions of this article are as follows:
\begin{itemize}
    \item For the privacy-preserving and device computing capability shortage problems in the IoT environment, we propose EUSFL, an edge-assisted U-shaped split federation learning framework, which effectively combines the distributed architecture of FL with SL. By splitting the model training process among the central server, edge server, and devices to achieve efficient distributed learning, and take full advantage of the data privacy-preserving features of FL. Moreover, a detailed performance analysis of EUSFL verifies its efficiency and stability in the IoT environment.
    \item We proposed an innovative noise mechanism named LabelDP, designed to safeguard client-held data against adversarial attacks. Through rigorous experiment, we assessed the extent of privacy enhancement facilitated by this algorithm.
    \item Extensive experiments were conducted to verify the effectiveness of the proposed EUSFL framework, and a detailed performance analysis was carried out. This is the first paper that provides both theoretical and empirical experiment about security and privacy in split learning and federated learning. The experimental results show that the EUSFL framework maintains good task accuracy over various FL algorithms. Furthermore, the EUSFL framework was also tested with different computational capabilities between devices and edge servers, and it was found that EUSFL effectively enhances the efficiency of performing deep learning model training in the IoT environment, significantly reducing the overall computing time. It was concluded that the EUSFL framework is applicable and scalable with good performance.
\end{itemize}

The rest of this article is organized as follows. In Section II, related research on edge intelligence, federated learning, and split learning is presented. Section III gives a brief knowledge description of federation learning and split learning. Section IV describes the overall content of the EUSFL framework and Section V gives a detailed performance analysis. Section VI presents the experiments and results analysis, and Section VII offers a conclusion.

\section{Related Work}

\subsection{Edge Intelligence}
With the rapid development of network communication technology and computing chips, IoT has begun to be widely deployed in people's production and living, which makes a large amount of data from different user devices need to be processed by various applications. Among them, machine learning technology provides a wide range of intelligent services, such as natural language processing, computer vision, and other applications that have greatly changed our lives.

Deep learning is one of the most commonly used AI methods to implement AI applications. Its principle is to put input data into a multi-layer neural network in the form of a vector and obtain the corresponding results through its calculation. The neural network can be regarded as a distribution function for a specific problem modelling. Its training process is to minimize the loss function by adjusting its parameters so as to fit this distribution function as much as possible, making it accurately perform the task we want. However, when deploying DL to various devices, a large amount of highly dispersed IoT data makes conventional cloud-based IoT infrastructure insufficient to support the original operation of applications. An emerging paradigm edge computing is to offload computing tasks to distributed cloud-like edge devices, promising a new infrastructure platform to support high-performance IoT applications.

Considering that existing applications have high computing requirements, although the performance of current end devices is gradually improving, it is still hard to meet the requirement for deep learning model training. Edge computing can be seen as an extension of cloud computing, which devolves computing, storage and other operations to edge nodes\cite{LearningIoTinEdge}. These edge nodes can be IoT gateways, routers, micro data centers, mobile base stations, etc.

This enables us to deploy AI algorithms and models directly on devices at the edge of the network rather than in the cloud or data center. Some examples of edge intelligence applications include car autopilot, industrial automation, and smart home devices \cite{edgeconfluence}. In this paper, our main purpose is to coordinate various edge intelligent devices in the context of the Internet of Things and edge computing to complete the training of neural networks, so as to provide more efficient and secure intelligent services.

\subsection{Federated Learning for Internet of Things}

FL is distributed machine learning technique that allows multiple devices, such as smartphones or edge devices train their models without sharing data. FL has greatly applied to various situations such as text and emoji prediction on smartphone keyboard\cite{keyboardpre}\cite{emojipre}, medical imaging\cite{medicalnature}, recommendation system\cite{recomendationsystem}, nature language processing\cite{fednlp}, and so on. This is implemented by having devices train their local models using their data, then sending model updates to the aggregator. The aggregator aggregates the updates and produces the updated global model. The popularity of federated learning keep growing in recent years due to its characteristic to protect user privacy, reduce data transfer, and support larger-scale training \cite{FLanalysis}.

In recent years, combining FL with IoT has become a hot research topic. The initial idea was to utilize the characteristics of federated learning to protect the data privacy of IoT users or devices. Haya et al. \cite{9509396} proposed a deep federated learning framework that uses IoT devices for healthcare data monitoring and analysis. Saurabh et al. \cite{singh2022framework} proposed an IoT medical data privacy protection framework that utilizes both federated learning and blockchain technology. As a new form of machine learning, FL also provides a new direction for IoT security issues. Thien et al.\cite{8884802} proposed a federated self-learning anomaly detection and prevention system that can detect and prevent unknown attacks in the IoT. Viraaji et al.\cite{9424138} proposed an anomaly detection method based on FL that uses decentralized device data in IoT networks to proactively identify intrusion attacks. In addition, FL also provides a new data-sharing mode for IoT environments, breaking down the data barriers of traditional IoT models. Lu et al.\cite{8843900} designed a secure data-sharing architecture for industrial IoT using blockchain technology and federated learning. FL has also been used for resource allocation in IoT environments. Nguyen et al.\cite{9187874} proposed a resource allocation problem that minimizes total energy consumption and FL completion time in wireless IoT networks using the FL algorithm. 

The combination of FL and edge intelligence can achieve real-time and decentralized machine learning at the edges, which is particularly useful for applications such as predictive maintenance, anomaly detection, and personalized recommendation systems\cite{FLChallenge}. However, applying FL to mobile edge networks faces many challenges, including data and system heterogeneity, communication overhead, limited resources, system reliability, scalability, and privacy and security \cite{FLChallenge}. Compared with common computing devices, the computing capability of IoT devices is usually weaker, which may cause the problem of straggling, affect the overall training time and the final model performance. Moreover, when edge devices train DL model locally, its high computation consumption may negatively affect the original task\cite{edgeFLSurvey}.

 For limited resources, existing research mainly considers reducing resource consumption by reducing data transmission. There are two major ideas to reduce transmission: (1) reduce the frequency of communication between clients and server (2) reduce the transmission between each client and server. For example, CMFL\cite{CMFL} reduces the overhead of data transfer by setting a threshold to reduce the upload of useless updates. In another method, \cite{FLCommOptim} reuses outdated local parameters to schedule clients and theoretically analyzed its communication characteristics and convergence. Although such methods can reduce communication overhead, FL cannot alleviate computational overhead.

\subsection{Split Learning}
Split learning is another form of distributed machine learning that allows a deep neural network to be split into at least two sub-networks trained on different distributed entities\cite{gupta2018distributed}. Initially, SL was proposed to address the predicament of data distribution and contributes to various applications, such as computer vision\cite{SLViT}, health care\cite{splitHealth}, sequential data\cite{SLRNN}, etc. Gupta et al.\cite{gupta2018distributed} and Vepakom et al.\cite{splitHealth} used the idea of SL to split DNN into multiple models, and then allowed multiple entities to train DNN with distributed data.

The separation mechanism of SL relieves the computing pressure of devices when training complex models to a certain extent, which is just suitable for resource-constrained IoT devices. Samikwa et al.\cite{samikwa2022ares} proposed Adaptive Resource-Aware Split Learning (ARES) for IoT environments, which enables efficient model training on resource-constrained IoT devices. However, on the other hand, SL also increases the communication consumption between devices. How to reduce the communication consumption of this process has become another research direction. Chen et al.\cite{ComReductionAsySL} proposed an asynchronous training scheme to reduce communication consumption by reducing the update frequency of the client model.  

In addition, as two different distributed machine learning methods, some studies have compared the two, or combined the two. Wu et al.\cite{SLparallel} and Gao et al.\cite{gao2020end} discussed the parallelism and resource management about SL, shows that the two have their own advantages and disadvantages in computing consumption and communication consumption. Thapa et al.\cite{SplitFed} proposed a new novel framework, combining FL and SL, used for model traning on distributed sequential data. Liu et al.\cite{SLUAV} applied such hybrid learning method to UAV cluster, greatly reduced computation overhead at end device. However, the author in \cite{fedorsplit} pointed out that the above training method is more effective for vertical federated learning where data and labels are dispersed. In the case of horizontal federated learning, spreading labels will cause privacy leakage.

In terms of privacy aspect, recent research has found that the smashed data also has potential to leak raw data information.




\section{Preliminaries}
In this section, we will describe some of the prior method in symbolic way, the symbols will be explained by 
Table \ref{Table_NotationDefinition}.

\begin{table}[tb]
\footnotesize
\centering
\caption{Symbols of this paper}
\tabcolsep 10pt 
    \begin{tabular*}{0.99\columnwidth}{cc}
    \toprule
    Symbols & Descriptions \\
    \midrule
    $F(W)$ & DL model \\
    $W$ & Model parameters\\
    $C_i^j$ & $i$-th client, affiliate to $j$-th edge server \\\
    $N$ & The number of Clients participate in training task \\
    $E_j$ & $j$-th Edge server \\
    $M$ & The number of Edge servers participate in training task \\
    $F_g$ & Up to date global model \\
    $F_1^i$ & Front split model for $i$-th client (first part) \\
    $F_2^i$ & Middle split model for $i$-th client (second part)\\
    $F_3^i$ & Rear split model for $i$-th client (third part)\\ 
    $L(\cdot)$ & Loss function \\
    $t$ & $t$-th training epoch \\
    $x$ & Data feature  \\
    $y$ & Data label, correspond to $x$ \\
    $d_F^{i}$ & smashed data generated by split model $F$ at $i$-th client \\
    \bottomrule
    \end{tabular*}
\label{Table_NotationDefinition}
\end{table}

\subsection{Federated Learning}
In the FL settings, $K$ participants cooperate with an aggregation server $A$ to train a global model $F(W)$. In each training epoch $t \in [1, T]$, the aggregation server randomly selects a subset $S
$ of participants, $|S|=r\cdot N \ge 1$ where $N$ is The total number of user participation, $r$ is the proportion of user selection. At the beginning of each round, the server $A$ distributes the current global model $W^t$ to the current round of participants $S$ and sends hyperparameters such as local epoch and batch size to the participants for training. After the participants' local training is finished, each participant uploads their respectively $W_k$ to the aggregation server for aggregation, as shown in the following equation:
\begin{equation}
    W^{t+1}=\sum_{k=1}^{K} \frac{d_{(k)}}{d} W_{(k)}^{t+1}
    \label{Eqa_FLDefinition}
\end{equation}

The FL optimization problem can be summarize as follow:  
\begin{equation}
\operatorname{minimize}_{W \in \mathbb{R}^{l}} \quad F(\mathbf{x}|W)=\frac{1}{D} \sum_{k=1}^{D} F_{k}(\mathbf{x}|W)
    \label{Eqa_FLOptim}
\end{equation}

\subsection{Split Learning}
Split learning splits an intact neural network $F$ into multiple parts, where $F = F_1 \cup F_2 \cup ...\cup F_n$. For any input data, there is $F(x) = F_1(F_2(...F_n(x)))$, which means the split models are equivalent to intact model. The training process is more complex than conventional deep learning because \textit{smashed data} will be transferred multiple times for forward propagation and backward propagation. In forward propagation, $F_i$ compute raw data $x$ to get smashed data $d_i$, and convey it to the holder of next split model, then keep computing $F_{i+1}(x)$ to get $d_{i+1}$ until the final part of split model $F_n(x) = output$. The final model holder compute loss value $l$, use $\nabla l$ to update model, then convey $\nabla d_{n-1}$ to last model holder, until the first part to split model $F_1$ get updated. The described steps is for single training epoch. To complete whole training process, these steps should be executed repeatedly too.

The split learning optimization problem can also be summarized as follow:
\begin{equation}
\mathop {argmin}_{W} \ L(F(x|W), y) \leftrightarrow \mathop {argmin}_{W} \ L(F_1...(F_n(x|W), y)
\label{Eqa_SLOptim}
\end{equation}

\subsection{Differential Privacy}

Differential privacy (DP) is a fundamental concept in the field of privacy-preserving data analysis. It provides a rigorous framework for quantifying and ensuring the privacy of individuals in datasets used for statistical analysis and machine learning. Introduced by Dwork et al.\cite{dwork2006differential}, differential privacy has become a cornerstone in the design of privacy-preserving algorithms and has gained significant attention from both academia and industry.

DP guarantees that the presence or absence of a single individual's data point does not significantly affect the outcome or result of any analysis. In other words, it ensures that the inclusion or exclusion of any individual's data does not lead to distinguishable changes in the output of queries or computations performed on the dataset. This property is achieved by adding carefully calibrated noise to the query results or the data itself, making it difficult to infer sensitive information about any individual, even if an adversary has auxiliary knowledge about the other data in the dataset.

A widely used algorithm is $(\epsilon, \delta)$-DP, which can be represented as follow:

\begin{equation}
    \operatorname{Pr}\left[M\left(D_{0}\right) \in S\right] \leq e^{\epsilon} \cdot \operatorname{Pr}\left[M\left(D_{1}\right) \in S\right]+\delta
\label{DP-definition}
\end{equation}

However, existing methods only add noises to data and parameters based on their application scenario. And for our framework, we use a novel LabelDP to protect our data against adversary. In LabelDP, we apply laplacian noise to labels before loss function calculation.

\begin{figure}[t]
\centering
\includegraphics[width=0.85\linewidth]{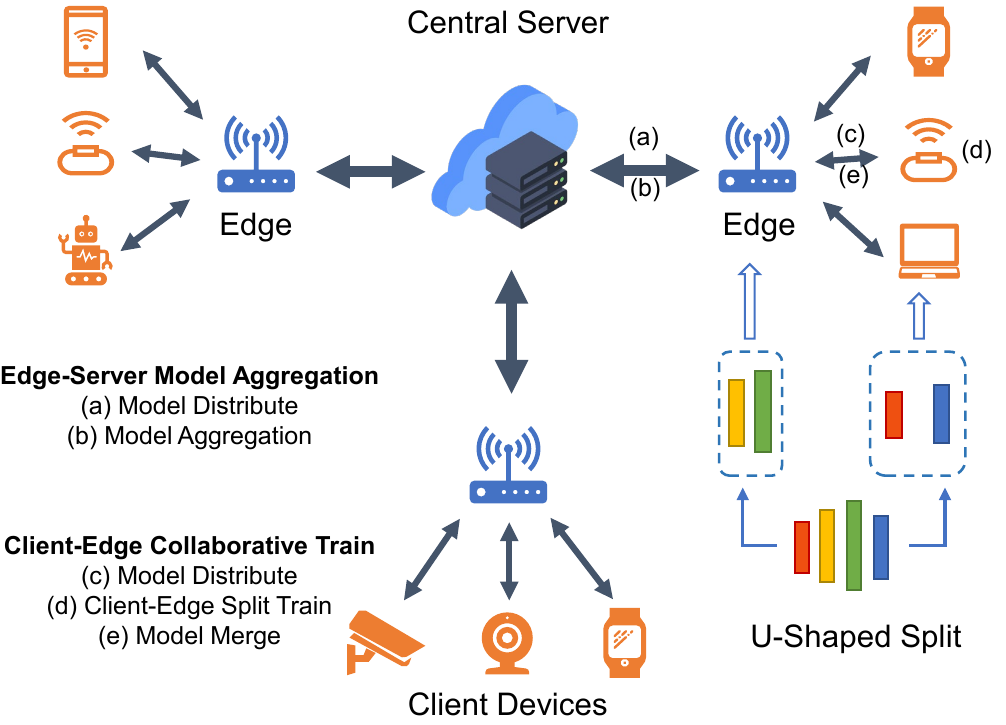}
\caption{EUSFL in IoT environment}
\label{Fig_IoT}
\end{figure}

\section{EUSFL Framework}
In this section, we give a comprehensive description and illustration of the EUSFL framework, including system design, U-shaped model split, and training process.
\subsection{System Design}

As shown in Fig. \ref{Fig_IoT}, we presented an edge-assisted three-tier IoT architecture, including three types of entities: \textbf{Central Server}, \textbf{Edge Server}, and \textbf{Client}. And we implement the server-edge-client collaborative training of models under this framework.

\textbf{Client}: Client is an end device deployed at the end of the network topology, with lightweight performance. It may correspond to the IoT device used to collect data in practical scenarios. The Client communicates with the Edge server or Central server, uploads and downloads various data. In the following sections, we use $C = \left\{c_1, c_2, ... ,c_N\right\}$ to denote $N$ participating Clients. Here $c_i$ represents the $i$-th For the participating Clients. The set of k Clients can be expressed as $C_k$ $(k<N) $.

\textbf{Edge Server}: The Edge server could be an upstream node of Client. Edge has higher computing and communication capabilities. In practical scenarios, it corresponds to high-performance devices that coordinate widely distributed IoT devices, such as workstations and IoT routers. In our scheme, Edge preserves the split model of the middle part corresponding to the Client. Compared with intact model, there are several vacant layers waiting for Client's aggregation, called \textit{Merge Layer}. At the end of each global epoch, the parameters uploaded by each Client are merged with the merge layers to obtain intact update model. The updated model will be uploaded to the Central server for aggregation.  In the next section, we use $E_m = \left\{ E_1, E_2,..., E_m \right\}$ to represent $m$ Edge servers $(m<<n)$, where $E_j$ denotes the $j$-th edge servers. For the $j$-th edge server, its subordinate client can be represented as $C_j$ , and $\sum_{1}^{j}C_j = |C|$ .

\textbf{Central Server}: The Server, stands on the top of the framework, has the highest computing and communication capabilities, and coordinates all the devices. In practical scenarios, it corresponds to the initiator of the entire task, usually a station's server. The Central Server is mainly responsible for receiving the intact models uploaded by Edge servers and aggregating the models before distributing them to each Edge before a new epoch. Besides, while owning the highest domination of the framework, the split mechanism $\mathcal{M}$ will be negotiated and distributed. In the following sections, $S$ will be used to denote the Central Server.

In FL, every Client performs local model training with the whole model, and upload model update to Central Server for aggregation. 

In vertical SL, the model is split into two parts, and they are placed at different entities. In this split method, along with smashed data the label data will be sent for loss function calculation. Compared with split learning, U-shaped split learning splits the model into three parts. The front and rear parts are placed at the Client, and the middle part is placed at the Edge server. This split method doesn’t require any data to be sent except smashed data.

Our method uses 3-layer architecture, combined with federated learning and U-shaped split learning. Between the Client and Edge server, the SL+FL method will be used for training and model aggregation, and between the Edge server and Central server, we will use FL model aggregation. 

\subsection{U-shaped Split}
\begin{figure*}[t]
\centering
\includegraphics[width=0.9\linewidth]{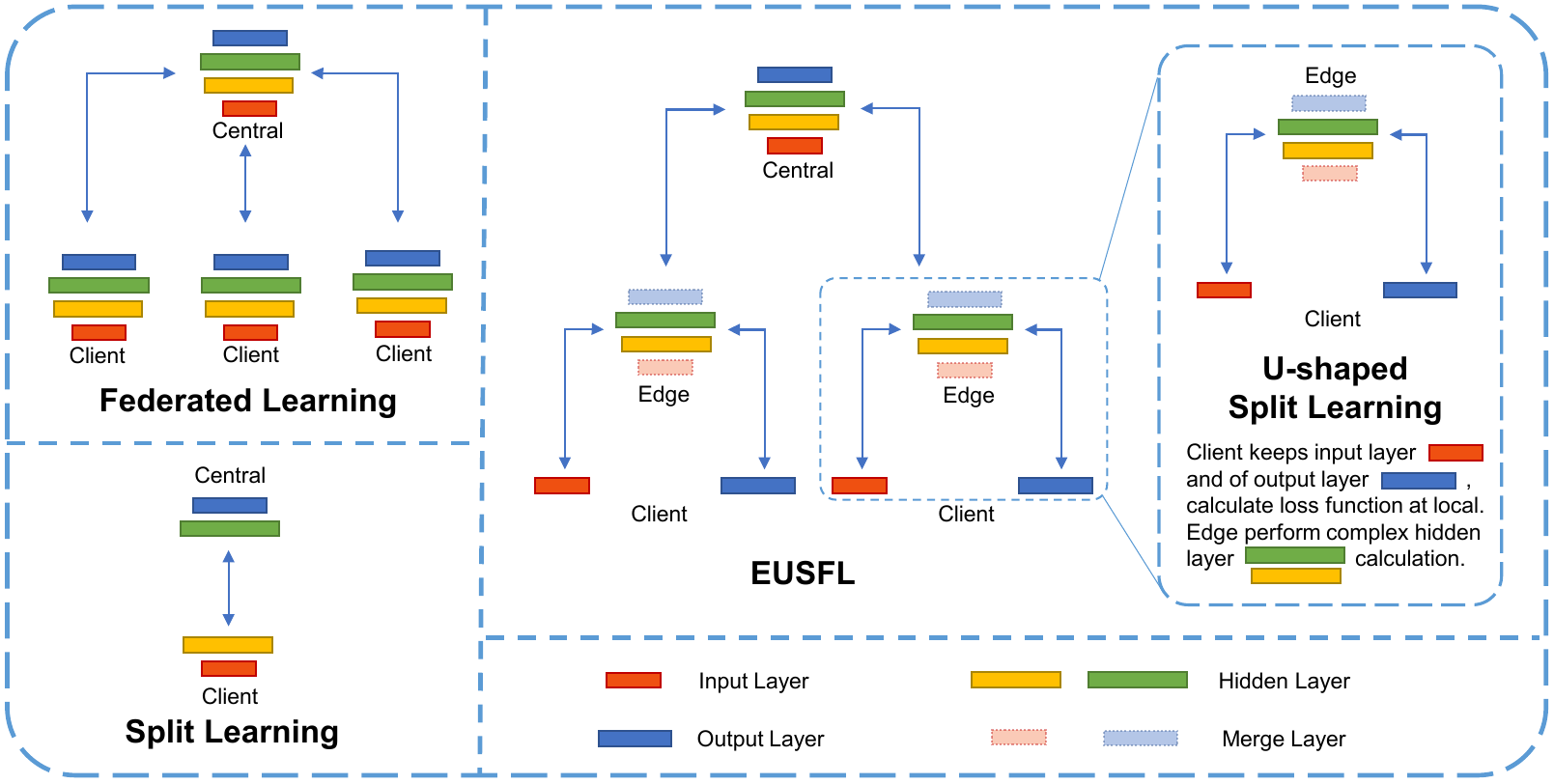}
\caption{U-shaped Split illustration and comparison}
\label{Fig_SystemDesign}
\end{figure*}
In our method, we assume each Client preserve entire data for training, rather than vertical training in which labels and data are dispersed. Therefore, taking into account the user's privacy issues, the label cannot be handed over for calculation. We use U-shaped split to solve this problem. As showed in Fig. \ref{Fig_SystemDesign}, the U-shaped structure split a intact neural network $F$ into three parts, namely the front part $F_1$, the middle part $F_2$ and the rear part, $F_3$, denotes as

\begin{equation}
F \leftrightarrow F_1 \cup F_2 \cup F_3
\label{Eqa_SLModel}
\end{equation}

Considering the neural network's data processing flow, a neural network is to input data into the input layer and uses the result obtained by the output layer, along with the label to calculate the loss function. For a single input data $x$, there is

\begin{equation}
    F(x) = F_3(F_2(F_1(x)))
    \label{Eqa_SLModelCal}
\end{equation}

This approach differs from the vertical split method, while U-shaped split enhanced data privacy. For comparison, we denote the vertical split method as SFL and the U-shaped split as USFL.

Therefore, for a complex network, the front and rear parts can be placed on the Client side for lightweight calculations, while the middle, complex parts can be placed on the Edge server side with higher performance. This ensures that the Client can cooperate with the Edge server to complete efficient model training without sharing any original data. For a more detailed approach, the U-shaped split can be seen from Algorithm \ref{Alg-U-split}.

\begin{algorithm}[t]
\footnotesize
  \SetAlgoLined
  \caption{U-shaped Split}
  \SetKwInput{Client}{$Client$}
  \SetKwInput{Edge}{$Edge$}
  \KwIn{$RawData = (x, y)$}
  \KwOut{Output data}
  \While{$i \le epoch$}{
    \Client{}
    \Indp 
    Compute $d_{F_1} \gets F_1(x)$\;
    Send $d_{F_1}$ to Edge Server\;
    \Indm
    \Edge{}
    \Indp 
    Compute $d_{F_2} \gets F_2(d_{F_1})$\;
    Send $d_{F_2}$ back to Client\; 
    \Indm
    \Client{}
    \Indp   
        Compute $predict \gets F_3(d_{F_2})$\;
        $ \text{Compute } Loss \gets L(predict, y) $\;
        $ \text{Compute } \nabla Loss \text{, Send to Edge server, Update } W_3 $\;
    \Indm
    \Edge{}
    \Indp 
       $\text{Update } W_2 \text{ ,Compute } \nabla d_{F_2} \text{ Send to Client} $\;
    \Indm
    \Client{}
    \Indp     
        $\text{Update } W_1$\;
    \Indm
  }
\label{Alg-U-split}
\end{algorithm}

\begin{table}[tb]
\footnotesize
\centering
\caption{Comparison on Different Schemes}
\tabcolsep 5pt 
    \begin{tabular*}{0.99\columnwidth}{lccc}
    \toprule
    \makecell{Models} & \makecell{Data \\ Protection} & \makecell{Label \\ Protection} & \makecell{Client Computation \\ Friendly} \\
    \midrule
    Distributed Learning\cite{distributedlearning} & × & × & \checkmark \\
    Federated Learning\cite{FL-source}   & \checkmark & \checkmark & × \\
    Split Learning\cite{splitHealth}       & \checkmark & × & \checkmark \\
    EUSFL                 & \checkmark & \checkmark & \checkmark \\
    \bottomrule
    \end{tabular*}
\label{Table_hhbh}
\end{table}

Table \ref{Table_hhbh} also gives a comparison of multiple learning schemes. For conventional distributed learning, the Client uploads all the data to a training center. Such method may cause privacy leakage and congestion, but the Client doesn't do any extra calculations. FL may bring calculation overhead because of the local computation on model training. Split learning can eliminate computation overhead to a certain extent, but conveying labels may also give rise to privacy problems. These methods have merits and drawbacks, combining these methods can eliminate the learning problem in the IoT environment.

\subsection{LabelDP}

For a laplacian distribution $Lap(x \mid \mu, b)=\frac{1}{2 b} e^{-\frac{|x-\mu|}{b}}$ used for adding noise, the parameter $\mu$ could be set to 0 to ensure the symmetry of the noise. Another parameter $b$ determines the noise scale, and we use sensitivity $\Delta f$ and privacy budget $\epsilon$ to describe it. In the following part, we use $Lap(\frac{\Delta f}{\epsilon})$ to represent a laplacian noise.

The process of LabelDP is to generate a $k$-dimension noise vector, then add to one hot label. That is, to generate a random vector $\Tilde{n} = (Lap_1(\frac{\Delta f}{\epsilon}), Lap_2(\frac{\Delta f}{\epsilon}), \dots , Lap_d(\frac{\Delta f}{\epsilon})), |n|=k$, then add to label $\mathbf{L}(y_i)$ to get $\Tilde{\mathbf{L}}(y_i)$ ($\mathbf{L}$ is one-hot function). And finally make it normalization like $\Tilde{\mathbf{L}}(y_i) = \frac{y_i}{\sum_{j=1}^{k} y_j}$. 

LabelDP makes training data noisy to resist reconstruction attack, the theoretical and empirical analysis will be presented at next chapter.

\subsection{Training Process}
In every global epoch, each client calculates the first part of the model (front part) and uploads the smashed data to Edge server. With stronger computing performance of the Edge server, Edge server can do its corresponding forward propagation with the second part model (middle part) quickly. Edge server uses smashed data to calculate the next smashed data and then sends it back to Client. Client uses the third part of the local model to calculate the prediction result, as well as loss value and gradient. Client Updates local parameters of the third part using gradient, and sends it back to the Edge server. The Edge server updates the parameters of the Edge server's middle model according to the result of the gradient and then sends it back to Client. Client finally updates the parameters of the first split model.

\subsubsection{Initialize}
Before initialization, we need the Central server, Edge server, and Client to negotiate a split mechanism $\mathcal{M}$ to ensure the split models are equivalent to the intact neural network. 
\begin{equation}
    \mathcal{M} : F_{E_j} \leftrightarrow F_1^{C_i} \cup F_2^{E_j} \cup F_3^{C_i}
    \label{Eqa_split_mechanism}
\end{equation}
In each epoch, the following steps will be executed in a loop until the target number of epochs is reached.

\subsubsection{Model Update}
The Central server sends the global model $F_G$ to Edge server, the Edge server updates the local model $F_E$ after receiving $F_G$ and loads the corresponding parameters into $F_2^{C_i}$. After that, Edge server broadcasts $F_{E}$ to its subordinate $C_j$, the subordinate clients also load the corresponding parameters into $F_1^{C_i}$ and $F_3^{C_i}$.

\subsubsection{Split Train}
During the process of training task, in each epoch, Client and Edge server need to jointly perform the following operations.

For forward propagation, for each data $TrainData (x, y)$ held by the Client, Client need to calculate $ F_1^i(x) \rightarrow d_{F_1}^i$ firstly using local model $F_1^i$, and the result $d_ {F_1}^i$ will be sent to its Edge server. Edge server uses the corresponding split model to get $ F_2^i(d_{F_1}^i) \rightarrow d_{F_2}^i $ , and sends the result $d_{ F_2}^i$ back to Client, Client than uses $F_3^i$ to calculate $F_3^i(d_{F_2}^i) \rightarrow output$.

For backward propagation, Client first calculates ${{\partial L(output,y)}/{\partial W_{3}^{i}}}$ to update the parameters of $W_3^i$, and then uses the same method to update $W_2^i$ and $W_1^i$. It is worth noting that in this process, $\nabla d_{F_2}^i$ needs to be uploaded to Edge server first, and then $\nabla d_{F_1}^i$ is downloaded to the Client after Edge server updates the parameters.

\subsubsection{Model Aggregation}
After Edge server and Client's models are all updated, the models need to be aggregated. First, the model merge will be performed between Client and Edge server. Clients upload the updated model parameters $W_1^{C_j}$ and $W_3^{C_j}$ to $E_j$, merge with $W_1^{C_j}$ and $W_3^{C_j}$ to $W_2 ^{E_j}$ to get $W_{E_j}$, and each Edge server uploads the merged new model $W_{E_j}$ to the Central server for second aggregation.

\begin{equation}
W_{g}^{t+1} = \frac{1}{|E|} \sum_{j=1}^{|E|}W_{E_j}
    \label{Eqa_train_aggregation}
\end{equation}

\begin{figure}[h]
\centering
\includegraphics[width=0.99\linewidth]{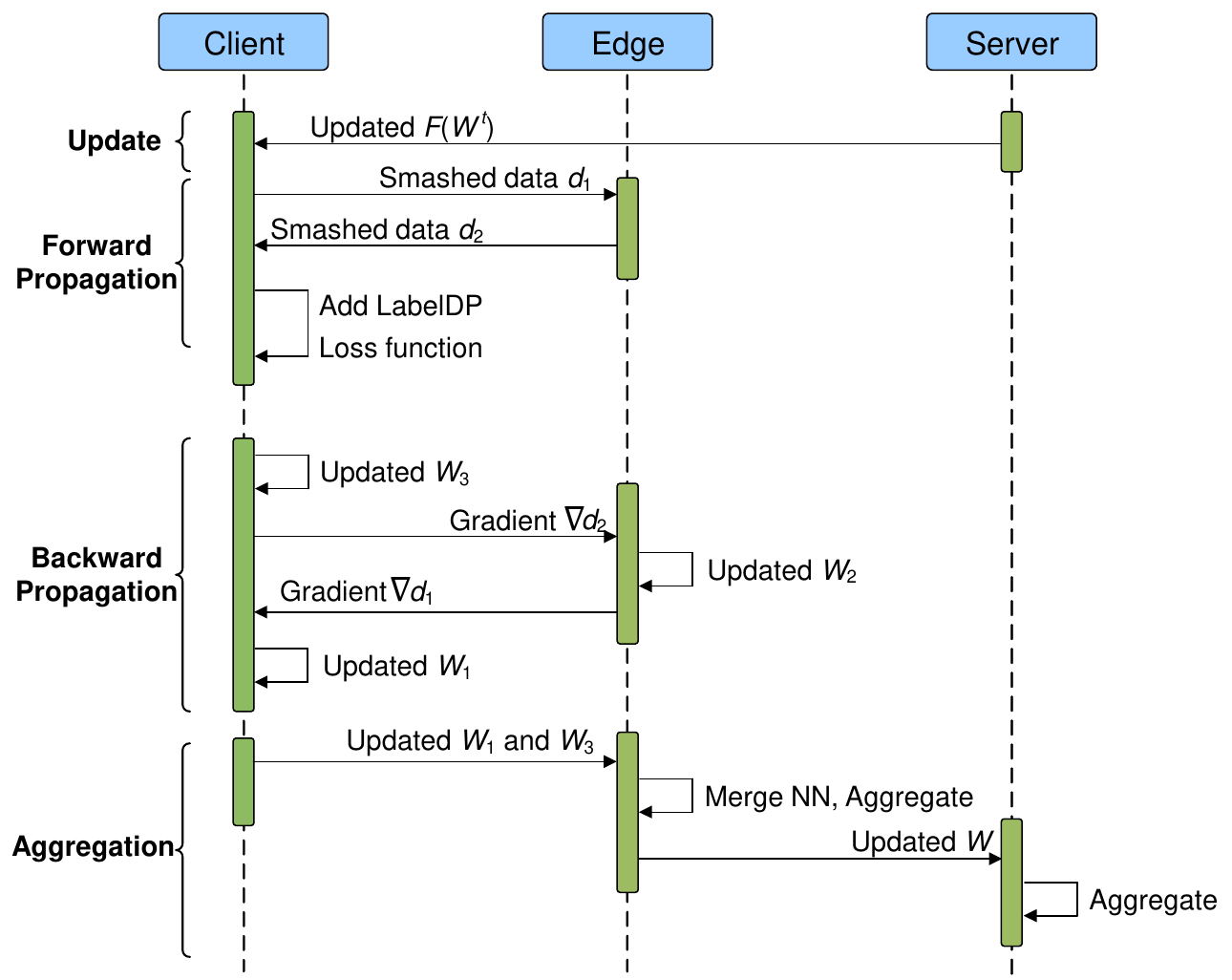}
\caption{Workflow of single epoch}
\label{Fig_workflow}
\end{figure}

After performing these steps, an epoch of training is finished. The overall training process can be seen from Algorithm \ref{Alg_overall}, Fig. \ref{Fig_workflow} also shows the workflow of the single epoch.

\begin{algorithm}[tb]
\footnotesize
  \SetAlgoLined
  \caption{Overall Training Process}
  \SetKwBlock{Initialize}{Initialize}{end}
  \SetKwBlock{Training}{Training}{end}
  \SetKwInput{Client}{$Client$}
  \SetKwInput{Edge}{$Edge$}
  \SetKwInput{Central}{$Central$}
  \KwIn{$RawData = (x, y)$}
  \KwOut{Final Global Model $F(W^t)$}
  \Initialize{
    \Central{Global Model $F(W^0)$\;}
    \Edge{Split Mechanism $\mathcal{M}$\;}
    \Client{Raw Data $d = (x,y)$\;}
  }
  \Training{
    \Central{$\text{Distribute } W^{t} \text{ to } E_j $\;}
    \While{$t \le epoch$}{
        \Edge{}
        \Indp     
            $\text{Split } W_1^t, W_2^t, W_3^t \gets \mathcal{M}(W^t) $\;
            $\text{Update } F_2^j \text{ Using } W_2^t$\;
            $\text{Distribute } W_1^t, W_3^t \text{ to } C_i^j$\;
        \Indm
        \For{$C_i^j \textbf{ in } C_j$}{
            \Client{}
            \Indp     
                $\text{Update } F_1^i, F_3^i \text{ Using } W_1^t, W_3^t $\;
                $\text{Compute } d_{F_1^i} \gets F_1^i(x) $\;
                $\text{Send } d_{F_1^i} \text{ to } E_j $\;
            \Indm
            \Edge{}
            \Indp     
                $\text{Compute } d_{F_2^j} \gets F_2^j(d_{F_1^i}) $\;
                $\text{Send } d_{F_2^j} \text{ back to } C_i^j $\;
            \Indm   
            \Client{}
            \Indp     
                $\text{Compute } predict \gets F_3^i(d_{F_2}) $\;
                $\text{Compute } y\prime \gets y + LabelDP $\;
                $\text{Compute } Loss \gets L(predict, y\prime) $\;
                $\text{Compute } \nabla Loss \text{ ,Send to } E_j $\;
                $\text{Update } W_3^{t+1} \text{ Using } \nabla Loss $\;
            \Indm
            \Edge{}
            \Indp     
                $\text{Update } W_2 \text{,Compute } \nabla d_{F_2} \text{,Send to } C_i^j $\;
            \Indm 
            \Client{}
            \Indp     
                $\text{Update } W_1$\;
                $\text{Upload } W_1^{t+1}, W_3^{t+1} \text{ to } E_j$\;
            \Indm
            \Edge{}
            \Indp     
                $\text{Merge } W_i^{t+1} \gets W_1^{t+1}\cup W_2^{t+1}\cup W_3^{t+1}$\;
                $\text{Compute } W_{j}^{t+1} = \frac{1}{|C_j|} \sum_{i=1}^{|C_j|}W_i^{t+1}$\;
                $\text{Upload } W_{j}^{t+1} \text{ to } S$\;
            \Indm 
            \Central{}
            \Indp     
                $\text{Compute } W^{t+1} = \frac{1}{|E_j|} \sum_{j=1}^{|E_j|}W_j^{t+1} $\;
            \Indm 
        }
    }
  }
\label{Alg_overall}
\end{algorithm}

\section{Efficiency Analysis}

The framework we propose optimizes both training efficiency and privacy enhancement. We will comprehensively analyze the advantages of our framework from these two perspectives.

\subsection{Training Time Analysis}

Considering the communication overhead brought by EUSFL, the time consumption may not be shorter than EFL. For example, in a scenario with low communication performance, Clients will spend much time waiting for the transmission of smashed data during the training process. Based on this condition, we make a detailed analysis of the efficiency of EUSFL.

Before analyzing the efficiency of our method, we will introduce other methods that we would like to compare, and these methods will also be conducted for experiments in next chapter.

If skipping the Edge server, the Client will directly connect to Central server. In this case, the transmission rate will be deficient, and it may cause connection interruption. In this way of connection, the most basic training method is FL - all clients directly connect to the Central server, and each client independently uses local data for training, only uploading model updates. The second basic training method is  SL - all clients place the shallow part of the neural network locally for data input and initial forward propagation and send smashed data to the Central server in each training epoch. After the Central server completes the loss function calculation, it sends the gradient of smashed data back to Client for backward propagation.

When using Edge server, the method mentioned in the previous paragraph will be combined with the 3-tier IoT architecture and extended to EFL, ESFL, and EUSFL. In this split method, Edge server will replace the Central server that is not applicable in the non-Edge method, and aggregation will be performed independently at both the Edge server and Central server.

Based on these conditions, the following section will show a comparison between the two methods as well as the derivation.

\begin{table}[tb]
\footnotesize
\centering
\caption{Efficiency Comparison among Various Methods}
\tabcolsep 3pt 
    \begin{tabular*}{0.99\columnwidth}{lccc}
    \toprule
    Methods & Edge Server  & Split Method & Time Consumption  \\
    \midrule
    FL      & ×          & \multirow{2}{*}{None}     &  \multirow{2}{*}{$\frac{2m}{r} + T_{c}$}  \\
    EFL     & \checkmark &                           &  \\
    SFL     & ×          & \multirow{2}{*}{Vertical} &  \multirow{2}{*}{$\frac{2m_1 + 2D_1\cdot|D_k|}{r} + T_{c} + T{e}$}  \\
    ESFL    & \checkmark &                           &  \\
    USFL   & ×          & \multirow{2}{*}{U-shaped} &  \multirow{2}{*}{$\frac{2(m_1 \! + \! m_3) + 2(D_1 \! + \! D_2)\cdot|D_k|}{r} \! + \! T_{c} \! + \! T_{e}$}  \\
    EUSFL  & \checkmark &                           &  \\ 
    \bottomrule
    \end{tabular*}
\label{Table_methods_analysis}
\end{table}

First of all, when using conventional FL, the FL process needs to go through three stages: (1)download updated model from Edge server and update local model. (2)local training. (3)model update(aggregation). So the time spent for FL to complete a single epoch is:
\begin{equation}
T_{epoch} = \frac{2m}{r} + T_c
    \label{Eqa_Tepoch}
\end{equation}

Where $m$ is the size of the model parameters, $T_c={FLOPs}/{P_c}$ refers to the time consumption of model inference, $FLOPs$ is the number of floating-point calculations required by forward propagation, and $P_c$ is the performance of Client (FLOPS), $r$ is the transfer rate to its upper entity.

If the model is split on the Client and Edge sides using U-shaped model split, the time required for USFL to finish a single epoch is:

\begin{equation}
\begin{split}
T_{epoch} =  & \frac{2(m_1+m_3) + 2(D_1+D_2)\cdot|D_k|}{r} \\ & + T_c + T_e
\end{split}
\label{deqn_ex1}
\end{equation}

Where $m_1, m_2, m_3$ are the sizes of $W_1, W_2, W_3$ (bytes), and there is $m=m_1+m_2+m_3$,  $D_1, D_2$ is the size of the smashed data $d_{F_1}, d_{F_2}$(bytes), $|D_k|$ is batch size of training. $T_c = {FLOPs_c}/{P_c}$ and $T_e = {FLOPs_e}/{P_e}$. $FLOPs_c$ is FLOPs required by Client to complete $F_1$ and $F_3$, $FLOPs_e$ is FLOPs required for the Edge server to complete the training of $F_2$. $P_e$ and $P_c$ is the performance of the Edge server and Client (FLOPS). There is $FLOPs = FLOPs_c + FLOPs_e$. 

In the same scenario, when the time consumption of merely using FL is longer than that of USFL, it is necessary to split the original FL.

That is to say when

\begin{equation}
\begin{split}
\Delta t = & (\frac{2m}{r} + \frac{Comp}{P_c}) -  (\frac{2(m_1+m_3)}{r} + T_c \\ & + T_e - \frac{2(D_1+D_2)}{r}\cdot|D_k|) > 0 
\end{split}
\label{deqn_ex1}
\end{equation}

The time spent training in USFL will be shorter than that in FL.

Given $m = m_1 + m_2 + m_3$, and $FLOPs = FLOPs_c + FLOPs_e$, the above equation can be simplified to get

\begin{equation}
\begin{split}
\Delta t = & \frac{FLOPs_e}{P_c} - \frac{FLOPs_e}{P_e} \\ & + (\frac{2m_2}{r}  - \frac{2(D_1+D_2)}{r})\cdot|D_k| >0
\label{deqn_ex1}
\end{split}
\end{equation}

The result of the above equation shows the time gap between FL and USFL in single epoch training, when the inequality is satisfied, USFL will be more efficient than FL. The larger the result is, the more efficient the USFL method will be. Adjusting the parameters of the above equation to make the result as large as possible will make USFL more efficient. This problem can be regarded as an multi-objective optimization problem, but the adjustment of the parameters may affect the accuracy of the training results, we will continue the this direction in our future work. For a more brief summary and analytical result, see Table \ref{Table_methods_analysis}.

\subsection{Privacy Analysis}


Consider the inherent privacy-preserving feature of FL is based on client's data preservation. In ideal circumstance, while all of the raw data is held by client, the privacy of these data is guaranteed. However, some researchers have found that exploiting alternative data could usurp privacy data. Besides, our method extends FL to more complex scenario, the privacy preservation capabilities should be discussed more meticulously.

\subsubsection{Threat Model}

Our ultimate objective is to train a federated model among massive IoT devices collaboratively and efficiently, exploiting privacy data held by clients while maintaining full data privacy. The potential threat could be as follow: Client to Client, Client to Edge server, and Client to Central server. 

There is Client to Client internal data stealing between clients, but this case usually happens in cross-silo scenario, conversely single participant in cross-device scenario is hard to distinguish. In IoT scenario, which is more similar to cross-device scenario, so we don't take such problem into consideration.

Another potential risk is about the Server, including Edge server and Central server. The Honest but Curious (HBC) adversary, which means this entity honestly execute its own task but curious try to obtain extra information, may try to reconstruct raw data using their accessible data. This aspect is the main concern that we take into consideration in our framework.

\subsubsection{LabelDP Analysis}

The following part will prove our LabelDP meets the criterion of $(\epsilon-\delta)$-Differential Privacy

For a original label $y$, and it's adjacent $y'$, with a randomize mechanism $\mathrm{R}$, and one-hot function $\mathbf{L}$, the difference of the two items can be $\Delta \mathrm{f}=\max _{y, y^{\prime}}\left\|\mathbf{L}(y)- \mathbf{L} \left(y^{\prime}\right)\right\|_{p}$. To add laplacian noise to $\mathbf{L}(y)$, the formatted representation could be as follow:

\begin{equation}
    \mathrm{R}(\mathbf{L}(y)) = \mathbf{L}(y) + (Lap_1(\frac{\Delta f}{\epsilon}), Lap_2(\frac{\Delta f}{\epsilon}), \dots , Lap_d(\frac{\Delta f}{\epsilon}))
\end{equation}

where $d$ is the dimension of $y$. 
 
Assume 
\begin{equation}
\begin{split}
    \mathbf{L}(y) &=\left(y_{1}, y_{2}, \ldots, y_{d}\right), \\
    \mathbf{L'}\left(y\right)& =\left(y_{1}^{\prime}, y_{2}^{\prime}, \ldots, y_{d}^{\prime}\right) \\ & = \left(y_{1}+\Delta y_{1}, y_{2}+\Delta y_{2}, \ldots, y_{d}+\Delta y_{d}\right)
\end{split}
\end{equation}

so $\Delta \mathrm{f}=\max \left(\sum_{i=1}^{d}\left|y_{i}-y_{y}^{\prime}\right|\right)=\max \left(\sum_{i=1}^{d}\left|\Delta y_{i}\right|\right)$

Assume all of the $y_i=0$, aforementioned $ \mathbf{L}(y)$ and $\mathbf{L}\left(y^{\prime}\right)$ can be simplified to $(0, 0, \dots, 0)$ and $ (\Delta y_1, \Delta y_2, \dots , \Delta y_d)$. Denotes the output of our randomize mechanism as $O=\left(o_{1}, o_{2}, \ldots, o_{d}\right)$, the probability of a specific result $O$ could be 

\begin{equation}
\begin{gathered}
\operatorname{Pr}[\mathrm{R}(\mathbf{L}(y_i))=O]=\prod_{i=1}^d \frac{\epsilon}{2 \Delta \mathrm{f}} e^{-\frac{\epsilon}{\Delta \mathrm{f}}\left|o_i\right|} \\
\operatorname{Pr}\left[\mathrm{R}\left(\mathbf{L}^{\prime}(y_i)\right)=O\right]=\prod_{i=1}^d \frac{\epsilon}{2 \Delta \mathrm{f}} e^{-\frac{\epsilon}{\Delta \mathrm{f}}\left|\Delta y_i-o_i\right|} \\
\end{gathered}
\label{LDP_final}
\end{equation}

Combine Equation.\ref{LDP_final} with the definition of $(\epsilon-\delta)$DP, the representation could be simplified as:

\begin{equation}
\begin{split}
\frac{\operatorname{Pr}[\mathrm{R}(\mathbf{L}(y_i))=O]}{\operatorname{Pr}\left[\mathrm{R}\left(\mathbf{L}^{\prime}(y_i)\right)=O\right]} = &\frac{\prod_{i=1}^d \frac{\epsilon}{2 \Delta \mathrm{f}} e^{-\frac{\epsilon}{\Delta \mathrm{f}}\left|o_i\right|}}{\prod_{i=1}^d \frac{\epsilon}{2 \Delta \mathrm{f}} e^{-\frac{\epsilon}{\Delta \mathrm{f}}\left|\Delta y_i-o_i\right|}} \\ =e^{\frac{\epsilon}{\Delta \mathrm{f}} \sum_{i=1}^d\left(\left|\Delta y_i-o_i\right|-\left|o_i\right|\right)}
\end{split}
\end{equation}

Consider

\begin{equation}
\sum_{i=1}^d\left(\left|\Delta y_i-o_i\right|-\left|o_i\right|\right) \leq \sum_{i=1}^d\left|\Delta y_i\right| \leq \max _{\mathbf{L}, \mathbf{L}^{\prime}}\left(\sum_{i=1}^d\left|\Delta y_i\right|\right)=\Delta \mathrm{f}
\end{equation}

The result could be as follow:

\begin{equation}
\frac{\operatorname{Pr}[\mathrm{R}(\mathbf{L}(y_i))=O]}{\operatorname{Pr}\left[\mathrm{R}\left(\mathbf{L}^{\prime}(y_i)\right)=O\right]} \leq e^{\varepsilon}
\end{equation}

The aforementioned process proved our LabelDP meets the criterion of $(\epsilon-\delta)$-Differential Privacy.

\subsubsection{Privacy Preservation Against HBC adversary}

\begin{figure}[h]
\centering
\includegraphics[width=0.99\linewidth]{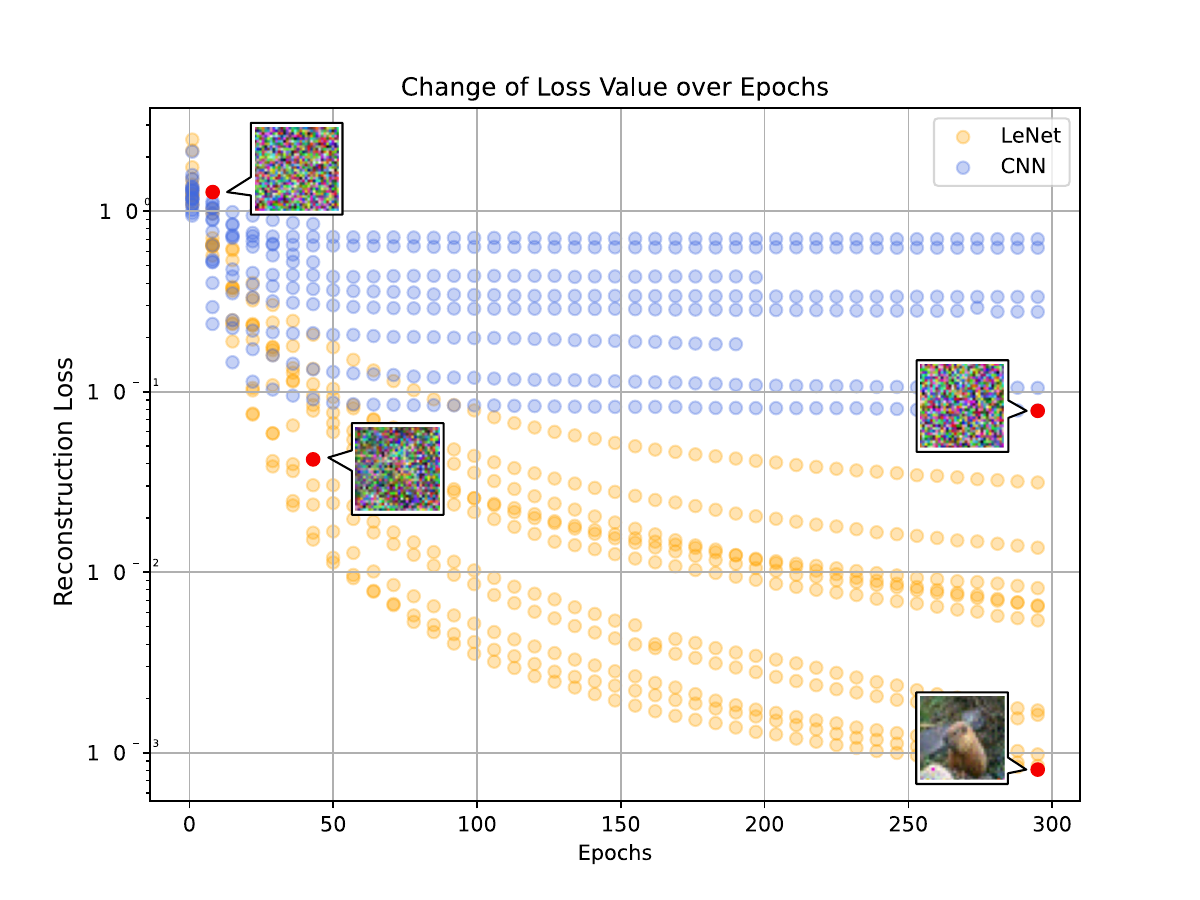}
\caption{Visualization of Image Reconstruction}
\label{Fig_dlg}
\end{figure}

There are several attacks could be used to recover raw data. DLG\cite{dlg} and iDLG\cite{idlg} is proposed to recover raw data and label, by optimizing the gradient difference between dummy data and ground truth. Although recovering data in such simple approach sounds sensational, there still exists convergence problem in deep models. Consider iDLG has better performace than DLG, we use iDLG to recover data. Fig.\ref{Fig_dlg} shows the result of iDLG implemented on Edge server side. The y-axis of this figure represents reconstruction loss, and lower loss indicates more accurate reconstruction. The yellow dots represent the original model (LeNet) implemented by the authors of iDLG, and blue dots represent our simplest model (CNN) in our experiment, which has more than about 10x parameters than LeNet. It can be observed that when the model becomes more complex, the reconstruction loss couldn't reach a visible extent.

\begin{figure}[h]
\centering
\includegraphics[width=0.99\linewidth]{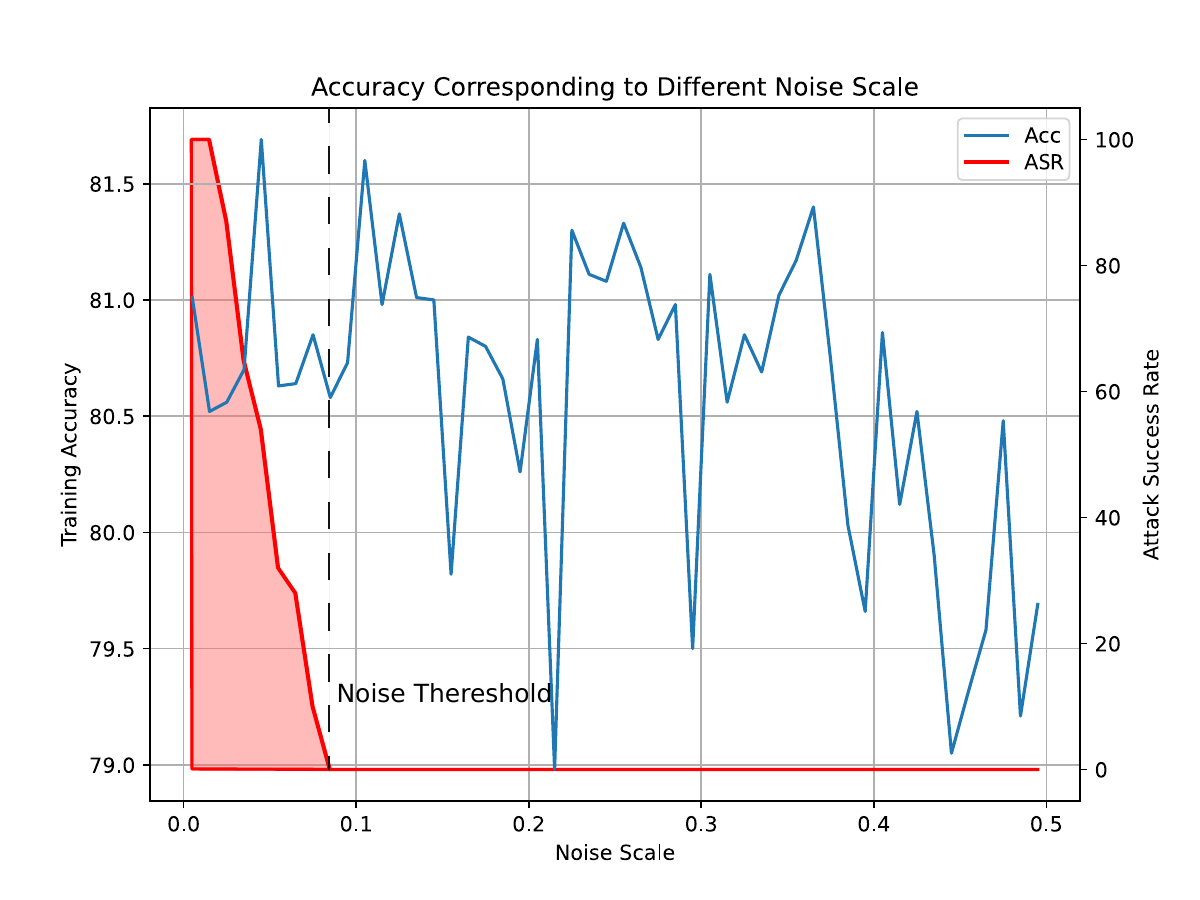}
\caption{Label Reconstruction Success Rate with Different Noise Scale}
\label{Fig_noise}
\end{figure}

However, reconstructing label data is a much easier task, our experiment shows without any additional protection, iDLG still works with 100\% successful rate even if parameters expand to 35M(VGG net). The advantage of our frame is that  

we use label-level noise to protect label data with additional LabelDP adds to label. Fig.\ref{Fig_noise} shows the the protection result of VGGNet under different noise scale. The left y-axis represents training accuracy after 10 epochs, and the right y-axis represents attack success rate (ASR). It can be observed that as the noise level increases, we can identify an optimal noise scale, which serves as the noise threshold. This threshold effectively resists label reconstruction attacks without accuracy deterioration.


FSHA\cite{FSHA} is another technique for non-data holders to reconstruct data. Instead of optimizing dummy data from gradients, this method uses a pivot model to train a reverse model that can revert raw input. However, the extensive preliminary process is too complex to conduct in an IoT environment. In this case, our framework can remain secure against such attacks.

\section{Experiment}
In this section, we describe the experimental settings used for evaluating the effectiveness and performance of EUSFL, including the comparison algorithm, and the datasets. Additionally, we present the experimental results and analysis.

\subsection{Experiment Setting}
This part describes the experimental environment, experimental setup, comparison methods, datasets, and metrics.

\subsubsection{Environment and Simulation Settings}
The experiment of EUSFL framework is conducted on a laptop with AMD Ryzen7 5800H, 16.0GB RAM, RTX 3060 6GB (Laptop version). The code of EUSFL and other compare methods are both based on Pytorch 1.8.2 version deep learning framework.

The computational and communication capabilities of IoT devices, Edge servers, and Central servers are simulated in the experiments, and the validation and performance evaluation of the method is performed with two different setups, as shown in Table \ref{Table_Eneities}.

For Setup 1 in practical scenario simulation, Clients have the computational performance of 400k FLOPS and transfer rate of 408kb/s to Edge servers. However, due to far distance and network topology constraints, the transfer rate to Server is only 20kb/s. The Edges have the computational performance of 8M FLOPS and transfer rate of 12Mb/s to the Server. The Server has a computational performance of 12M FLOPS. 

For Setup 2 in practical scenario simulation, Clients have the computational performance of 1G FLOPS and transfer rate of 8Mb/s to Edge servers. And due to those constraints, the transfer rate to Central server is 100kb/s. Edge servers have the computational performance of 20G FLOPS and transfer rate of 12Mb/s to Central server, and the Central server has a computational performance of 30G FLOPS.

\subsubsection{Datasets and Comparison Methods}

\begin{table}[h]
\footnotesize
\centering
\caption{Simulation Setup for Different Entities}
\tabcolsep 10pt 
    \begin{tabular*}{0.99\columnwidth}{lccc}
    \toprule
              Entity              &      Action                     & Setup 1 & Setup 2 \\ 
    \midrule
    \multirow{3}{*}{Client} & Computation               & 400k    & 1G      \\
                            & Communication with Edge   & 408kb/s & 8MB/s   \\
                            & Communication with Central & 20kb/s  & 100kb/s \\
    \midrule
    \multirow{2}{*}{Edge}   & Computation               & 8M      & 20G     \\
                            & Communication with Central & 12MB/s  & 12MB/s  \\
    \midrule
    Central                  & Computation               & 12M     & 30G     \\
    \bottomrule
    \end{tabular*}
\label{Table_Eneities}
\end{table}

There are two datasets in our experiment, including MNIST and CIFAR-10.

\textbf{MNIST} dataset consists of 70,000 grayscale images, with a size of 28*28 pixels. There are 60,000 images in the training set and 10,000 images in the testing set. Every image represents a corresponding digit from 0 to 9. The total size of this dataset is about 52.4MB.

\textbf{CIFAR-10} consists of 60,000 32x32 color images in 10 classes, with 6,000 images per class. The dataset is divided into 50,000 training images and 10,000 testing images. The total size of this dataset is about 175.78MB.

\begin{figure*}[]
\centering
\includegraphics[width=\linewidth]{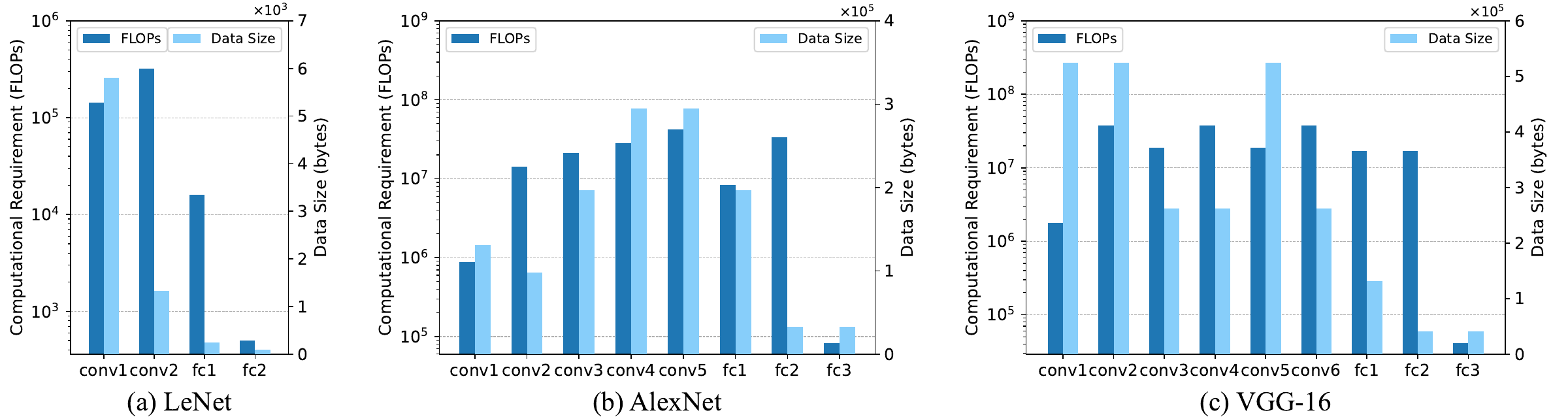}
\caption{Details of neural network model split}
\label{Fig_models}   
\end{figure*}

Simultaneously, we use \textbf{LeNet}, \textbf{AlexNet}, and \textbf{VGG-16} to process these data. For MNIST dataset, we use LeNet to process it, which need 30.75M FLOPs to train a batch with 64 images. The model parameters of LeNet is about 22k. For CIFAR-10 dataset, we use AlexNet and VGG-16 to process. AlexNet has about 23.27M model parameters, and 9.46B FLOPs to finish a 64 images batch training. VGG-16 has about 34.75M model parameters and 11.98B FLOPs to train a 64 images batch. Additionally, detailed FLOPs and Data Size information of the neural network models are presented in Fig.\ref{Fig_models}. 

For the accuracy comparison of EUSFL combined with different methods (\ref{accuracy comparison}), we use the following FL algorithms: 

\begin{itemize}
    \item 
    \textbf{FedAvg}\cite{FL-source} uses multiple local updates and aggregates models by averaging.
    \item 
    \textbf{FedProx}\cite{FedProx} mitigated the problem of imbalanced data distribution by adding a proximal term to the optimization objective, which attaches local model weights more than closer to the global model weights during the update.
     \item 
    \textbf{Scaffold}\cite{Scaffold} is an approach to mitigate communication by preserving a control variable. This approach mainly focused on gradient sparsity, making communication between devices more compact.
    \item 
    \textbf{FedNova}\cite{FedNova} normalizes the local model parameters, and the Server also collects normalizes, and averages the updates from all devices to reduce inconsistency.
    \item 
    \textbf{FedDC}\cite{FedDC} uses drift variables to track local parameter drift between the local and global models.
\end{itemize}

For better understanding and illustration, the comparison methods that appear in the experiments to evaluate EUSFL performance are summarized as follows.
\begin{itemize}
    \item \textbf{FL:} represents the original Federated Learning approach without any operational improvements, and the clients interact with the Central server directly.
    \item \textbf{EFL:} FL framework with Edge server as an intermediate layer, where the Edge server interacts between the Central server and the clients.
    \item \textbf{SFL:} Split Federated Learning, no Edge server as the intermediate layer, using vertical split method, only simply dividing the neural network model into two parts for execution, without considering the risk of privacy leakage of label information.
    \item \textbf{USFL:} U-shaped split Federated Learning, no Edge server as the intermediate layer, using U-shaped split method, divides the neural network model into three parts, and the middle part of the model is computed in the central server, while the other parts are computed in the client, without label information transmission.
    \item \textbf{ESFL:} Edge-assisted split Federated Learning, with the Edge server as the intermediate layer and vertical split method for the neural network model, where the client performs the first part of the neural network model and the Edge server performs the other parts, which has the risk of label information leakage.
    \item \textbf{EUSFL:} Edge-assisted U-Shaped split Federated Learning, with the Edge server as the intermediate layer and U-shaped split for the neural network model, the Edge server executes the middle part of the neural network model, the client executes the other parts, keeping the label information local to the client, without any label information leakage.
\end{itemize}

\subsubsection{Metrics}
In terms of computational performance, we use FLOPS, the number of floating-point operations per second to measure the performance of an entity. The operation of floating point numbers includes basic addition, subtraction, multiplication, and division. For example, when multiplying the $m*n$ dimensional matrix $A$ and the $n$ dimension vector, each number of the matrix will be multiplied with each number of the vector, and then add them all to get the sum of this column. In this process, $2n$ floating-point operations are performed, so when calculating the result, $2mn$ floating-point operations will be performed. When the neural network executes forward propagation, the input data needs to be multiplied by the matrix composed of the weight and bias of each layer, and the result is conveyed to the next layer for progressive calculation, until the network output the prediction value.

To measure inference time for a model, we can calculate the total number of computations the model will have to perform. This is where we mention the term FLOPs or Floating Point Operation. The FLOPs will give us the complexity of our model.

In the following discussion, the time required to perform single epoch model training can be obtained by dividing FLOPS by the number of floating-point number operations that the model can perform per second. That is $t = {FLOPs}/{FLOPS}$.

According to the description in the last section, predicting an image can be considered as a classification problem, which converts a 28*28-dimension or 3*32*32-dimension vector to a 10-dimension prediction vector. The performance of the model is mainly tested by the accuracy of the model, $Acc = {Test Corrects}/{TotalTest} *100\%$, since SL will not have any influence on the accuracy of the model, in the comparison of EFL, ESFL, and EUSFL, we mainly use the training time to compare.

\subsection{Result and Analysis}

In this section, we discuss the feasibility, and scalability of the EUSFL framework, and illustrate three key factors that affect training efficiency, including Client's computational performance, the Client's transfer rate to Edge server, and Edge server's computational performance. Finally, there are two practical simulations to show our scheme can greatly shorten the process of training.

\subsubsection{Accuracy and Scalability}
\label{accuracy comparison}
Before discussing training efficiency, we conduct some experiments to verify there is no effect of EUSFL on training accuracy. In this part, the ESFL and EUSFL methods are combined and applied with five different FL algorithms, which compare the accuracy with different datasets and neural network models with epoch settings of 10, 20, and 30, respectively. As shown in Table\ref{Table_training_acc}, in some scenarios, EUSFL has better accuracy than the original FL methods without the EUSFL framework. As an example, in the epoch of 10, MNIST dataset experiments, EUSFL when combined with FedProx, Scaffold, FedNova, and FedDC all have better accuracy than those without the EUSFL framework. Similarly, the performance is almost similar in CIFAR-10 datasets with other epoch settings.

\begin{table*}[tb]
\footnotesize
\centering
\renewcommand\arraystretch{0.6}
\caption{Accuracy Comparison on EUSFL Combing Different FL Methods}
\tabcolsep 9pt 
    \begin{tabular*}{0.99\linewidth}{lcccccccccccc}
    \toprule
    \multirow{2}{*}{Training Method}                & \multicolumn{3}{c}{MNIST + LeNet} & \multicolumn{3}{c}{CIFAR-10 + LeNet} & \multicolumn{3}{c}{CIFAR-10 + AlexNet} & \multicolumn{3}{c}{CIFAR-10 + VGG} \\
    \cmidrule(l{2mm}r{2mm}){2-4} \cmidrule(l{2mm}r{2mm}){5-7} \cmidrule(l{2mm}r{2mm}){8-10} \cmidrule(l{2mm}r{2mm}){11-13}
              & 10        & 30        & 50        & 10         & 30         & 50         & 10          & 30          & 50         & 10         & 30        & 50        \\
    \midrule
    FedAvg               & 94.55     & 97.41     & 98.02     & 28.31      & 44.84      & 52.81      & 24.48       & 58.77       & 70.74      & 59.36      & 76.96     & 81.32     \\
    ESFL+FedAvg              & 94.73     & 97.45     & 98.10     & 26.81      & 45.96      & 52.73      & 21.78       & 58.69       & 70.81      & 61.31      & 76.24     & 81.35     \\
    EUSFL+FedAvg            & 91.70     & 97.65     & 97.86     & 27.71      & 45.72      & 52.82      & 23.73       & 58.21       & 70.88      & 59.51      & 77.36     & 81.34     \\
    \midrule
    FedProx     & 85.01     & 91.96     & 97.82     & 20.82      & 40.16      & 51.78      & 16.36       & 54.57       & 69.55      & 50.96      & 72.28     & 80.20     \\
    ESFL+FedProx    & 83.36     & 92.92     & 97.92     & 21.47      & 40.19      & 51.60      & 16.01       & 55.05       & 69.40      & 51.71      & 71.92     & 80.01     \\
    EUSFL+FedProx  & 86.06     & 92.20     & 97.75     & 21.22      & 39.86      & 51.80      & 15.96       & 55.65       & 69.37      & 51.96      & 71.53     & 80.07     \\
    \midrule
    Scaffold    & 87.25     & 93.96     & 97.84     & 20.56      & 41.24      & 51.37      & 17.03       & 54.90       & 69.17      & 53.76      & 73.15     & 79.84     \\
    ESFL+Scaffold   & 87.05     & 93.96     & 98.01     & 20.56      & 41.84      & 51.42      & 17.68       & 55.08       & 69.02      & 54.51      & 73.00     & 79.71     \\
    EUSFL+Scaffold & 87.95     & 93.72     & 97.87     & 20.66      & 40.64      & 51.53      & 17.98       & 55.26       & 69.05      & 55.26      & 72.73     & 79.55     \\
    \midrule
    FedNova     & 93.29     & 94.37     & 97.91     & 22.37      & 42.20      & 52.40      & 21.69       & 55.57       & 68.76      & 57.20      & 74.68     & 80.45     \\
    ESFL+FedNova    & 94.14     & 94.55     & 97.75     & 21.92      & 41.84      & 52.26      & 22.39       & 55.03       & 68.66      & 57.70      & 74.38     & 80.51     \\
    EUSFL+FedNova  & 94.74     & 94.22     & 97.64     & 21.22      & 42.08      & 52.30      & 22.24       & 54.73       & 68.67      & 58.00      & 74.23     & 80.36     \\
    \midrule
    FedDC       & 88.80     & 91.71     & 97.70     & 22.86      & 40.97      & 51.19      & 19.18       & 53.94       & 69.68      & 52.51      & 72.73     & 80.40     \\
    ESFL+FedDC      & 89.65     & 91.47     & 98.04     & 23.41      & 41.42      & 51.07      & 19.58       & 54.03       & 69.57      & 53.36      & 72.52     & 80.31     \\
    EUSFL+FedDC    & 89.55     & 91.41     & 97.57     & 23.36      & 41.39      & 51.12      & 19.08       & 54.42       & 69.52      & 52.36      & 72.67     & 80.13     \\
    \bottomrule
    \end{tabular*}
\label{Table_training_acc}
\end{table*}

\begin{figure}[tb]
\centering
\subfigure[MNIST]{
\includegraphics[width=0.475\linewidth]{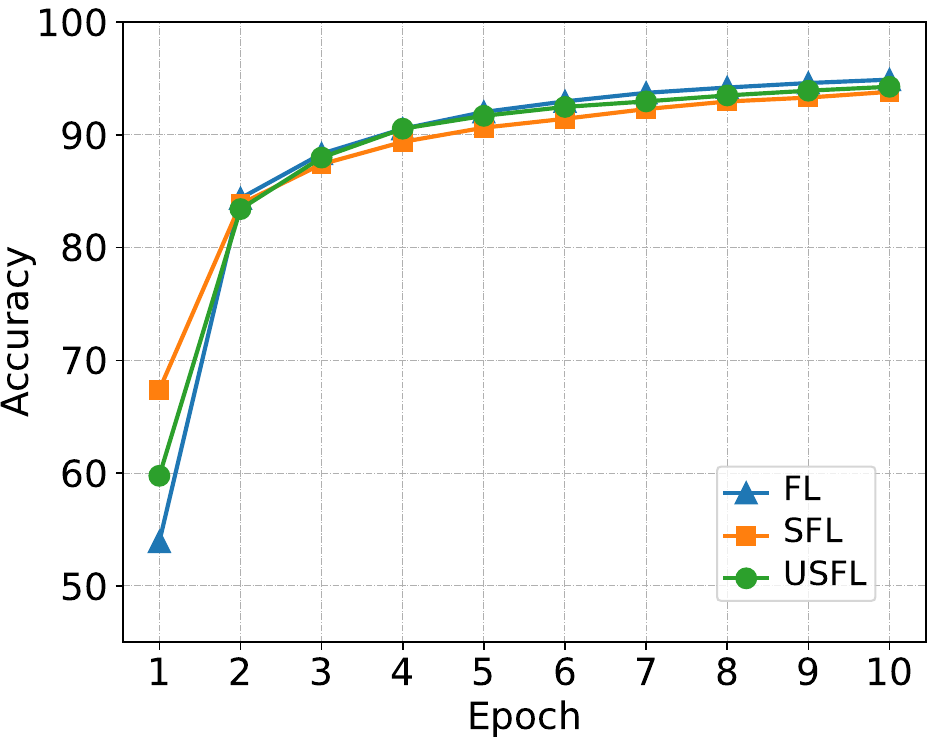}
\includegraphics[width=0.475\linewidth]{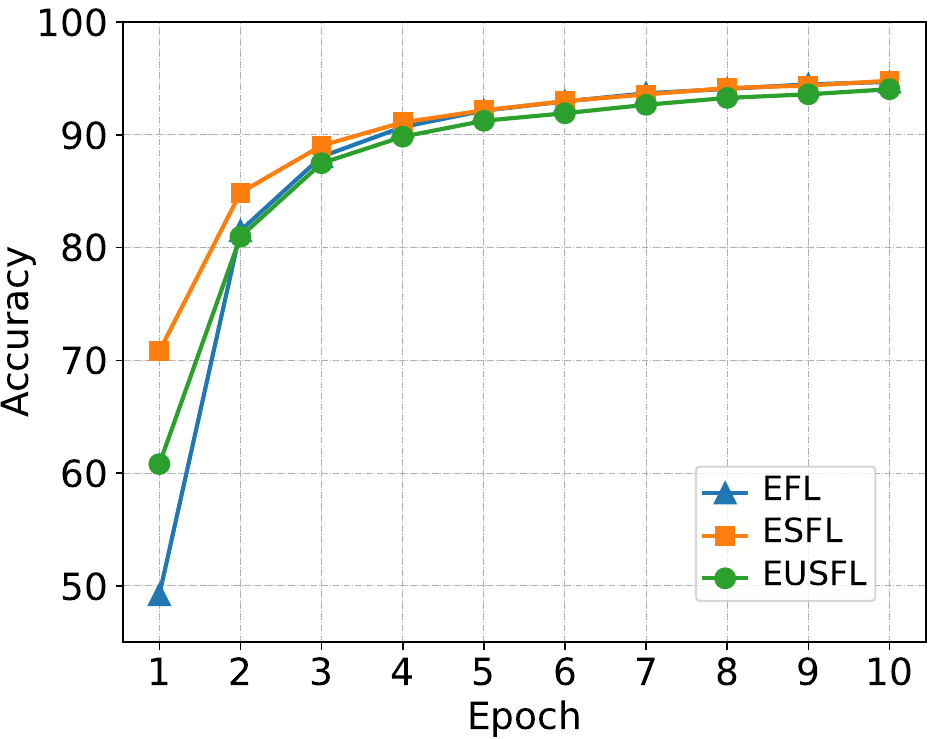}
}
\hfil
\subfigure[CIFAR-10]{
\includegraphics[width=0.475\linewidth]{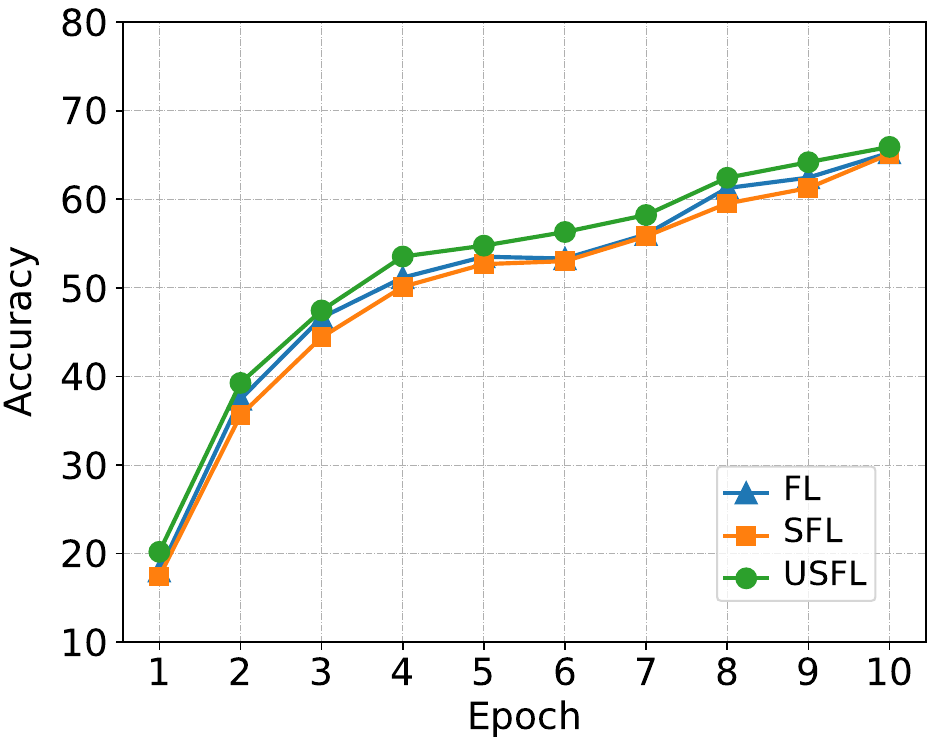}
\includegraphics[width=0.475\linewidth]{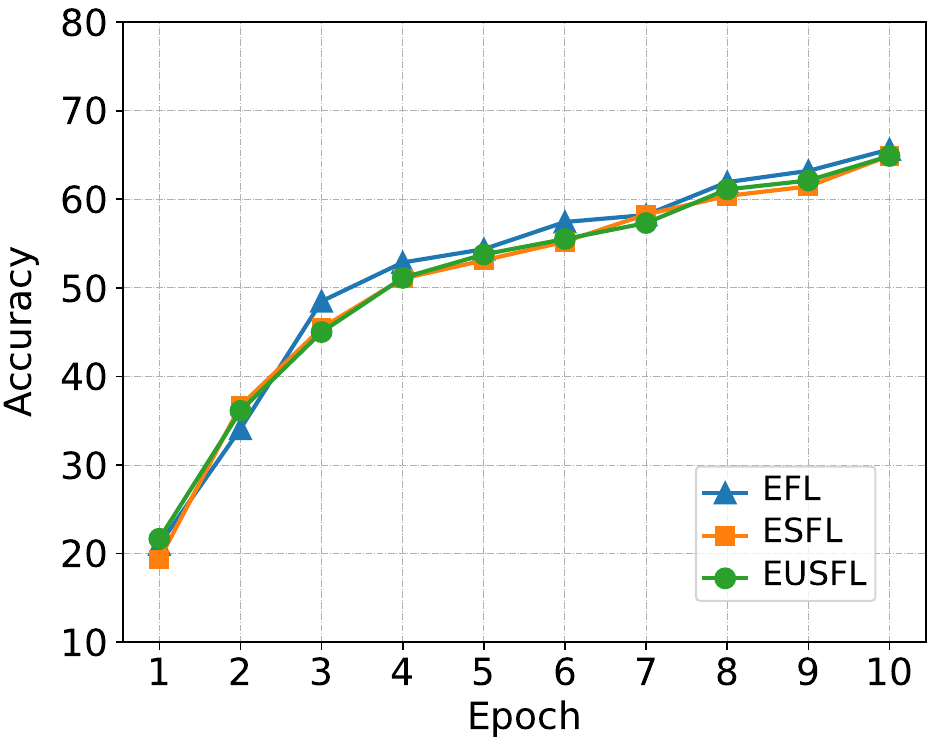}
}
\caption{Accuracy after 10 epochs on various methods}
\label{Fig_mnist_vgg_acc}
\end{figure}

Fig. \ref{Fig_mnist_vgg_acc} shows the accuracy of LeNet on the MNIST dataset and VGG-16 on the CIFAR-10 dataset using various training methods after 10 epochs. It can be observed that there is little difference in accuracy between the different methods after training, as the overall training process of both standard SL and U-shaped SL is mathematically equivalent to that of FL. However, U-shaped SL offers better privacy and does not involve the transmission of label information during the training process, thus enhancing its privacy-preserving capability.

Since these methods are mathematically equivalent while model training, in the following part we will focus on training efficiency, and we use the time consumption for finishing the training task to measure the efficiency of the methods in the same settings. 

\subsubsection{Computing Performance}

\begin{figure}
    \centering
    \includegraphics[width=0.85\linewidth]{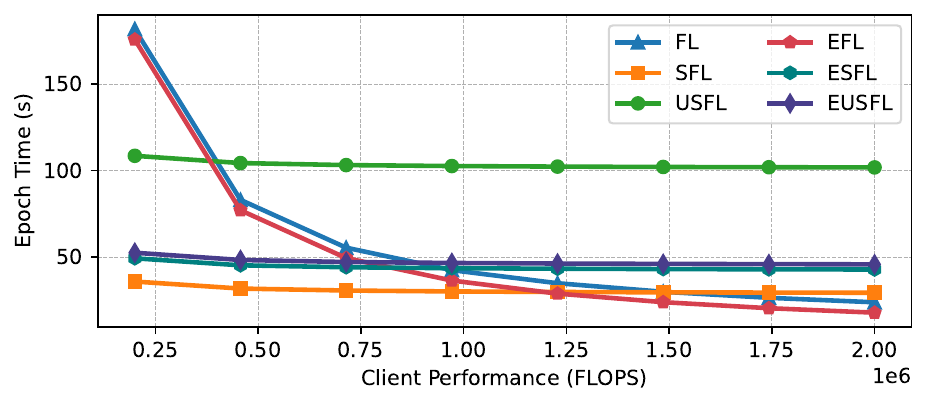}
    \caption{Client Performance Comparison}
    \label{Fig_client_computation}   
\end{figure}

Fig. \ref{Fig_client_computation} presents the effect of Client performance on single-epoch training, with all other variables held constant. As shown in the figure, the time required for single-epoch training decreases as Client performance improves. Notably, FL and EFL are the methods most significantly affected by performance, as these two methods do not split the neural network, and the training is entirely performed on the Client side. Thus, if the Client's computing performance is sufficient, FL's training speed is relatively fast.

However, for cases where Client performance is insufficient, single-epoch training using split methods takes less time than FL. This finding reinforces the conclusions presented in the performance analysis chapter.

When Edge server usage is absent, like SFL and USFL, suffer from longer completion time. This is due to the direct communication between Clients and the Central server is relatively inefficient, and the high communication demands imposed by split learning amplify this issue. As a result, even with improved Client performance, the time it takes to complete single-epoch training remains extensive.

Among all the methods, using SFL directly results in the shortest completion time. However, SFL, not utilizing the U-shaped split setting, requires label delivery, which raises privacy concerns that need to be addressed before implementation. Therefore, this method is not being considered.

\begin{figure}[!t]
\centering
\includegraphics[width=0.85\linewidth]{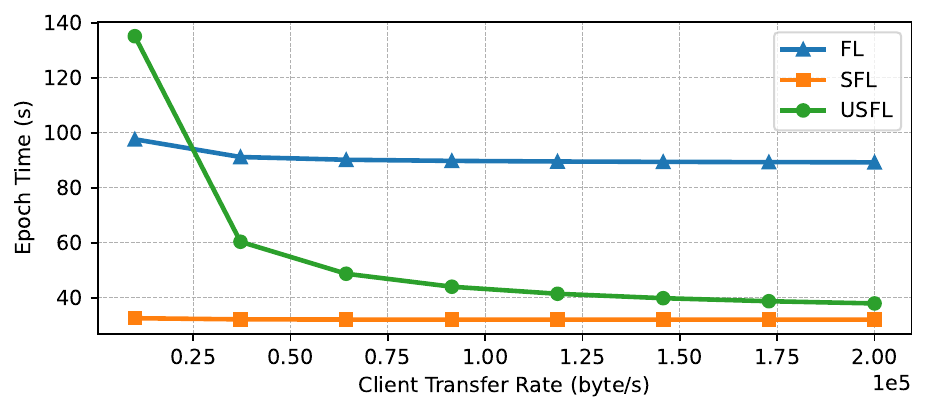}
\caption{Client Transfer Rate Comparison}
\label{Fig_client_transfer}   
\end{figure}

\subsubsection{Communication}

Fig. \ref{Fig_client_transfer} shows the influence of Client transfer rate on single-epoch training time, with all other variables held constant except for the transfer rate between the Client and its upper entity, including the Central server or Edge server. 

In this experiment, the use of Edge server is not included, and only three methods are compared. Since the use of Edge server primarily affects the Client transfer rate, when the transfer rate is slow, it can be regarded as the Client is directly connected to the Central server, i.e. FL, SFL, and USFL. On the other hand, when the transfer rate is high enough, the Client is connected to the Central server via Edge server as an intermediate node, i.e. EFL, ESFL, and EUSFL. 

The result shows U-shaped split is the method most affected by transfer rate, as U-shaped split has the highest communication requirements. Among these methods, SFL has the lowest communication requirements, as it only needs to transmit intermediate data and front part model, which is much smaller than model parameters in FL, so an increase in transfer rate does not have a significant impact on its single epoch training time.

The thing to note is that, although we are not considering the SFL method due to privacy concerns, USFL can be more efficient than FL when certain communication capacity requirements are met.

\subsubsection{Edge Server Performance}

\begin{figure*}[]
\centering
\subfigure[EFL]{\includegraphics[width=0.25\linewidth]{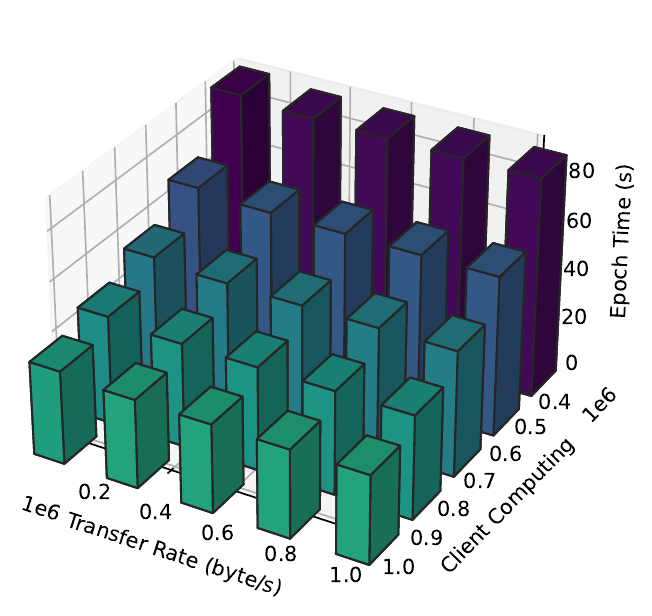}}
\hfil
\subfigure[ESFL]{\includegraphics[width=0.25\linewidth]{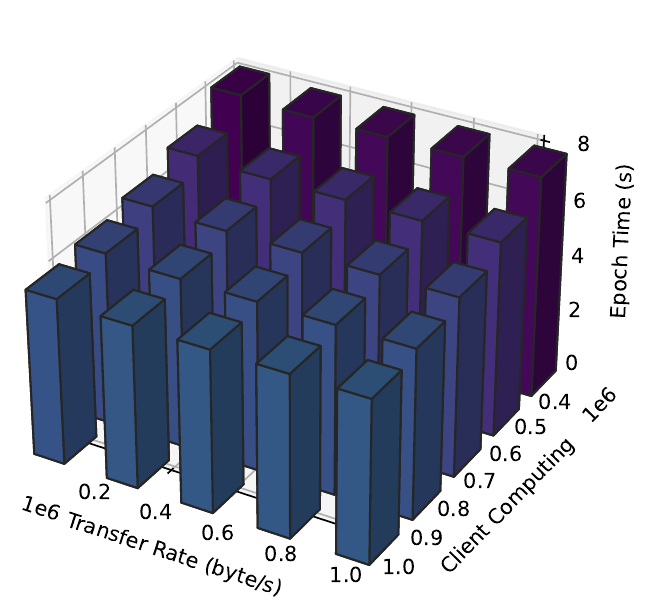}}
\hfil
\subfigure[EUSFL]{\includegraphics[width=0.25\linewidth]{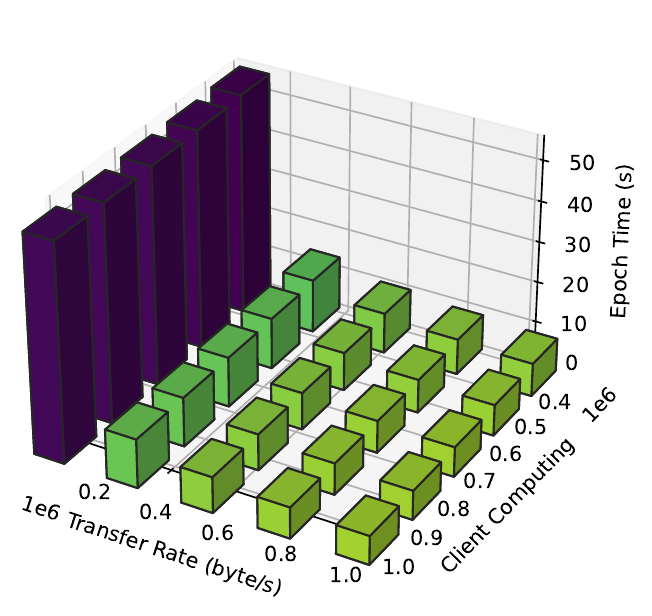}}
\caption{Influence of Client Performance and Transfer Rate on Single-epoch Training}
\label{Fig_client_3d}   
\end{figure*}

However, in practice, the computing performance of the Edge server is also a crucial factor to consider because, during split training, part of the neural network needs to be computed by Edge server. 

\begin{figure}
    \centering
    \subfigure[]{\includegraphics[width=0.475\linewidth]{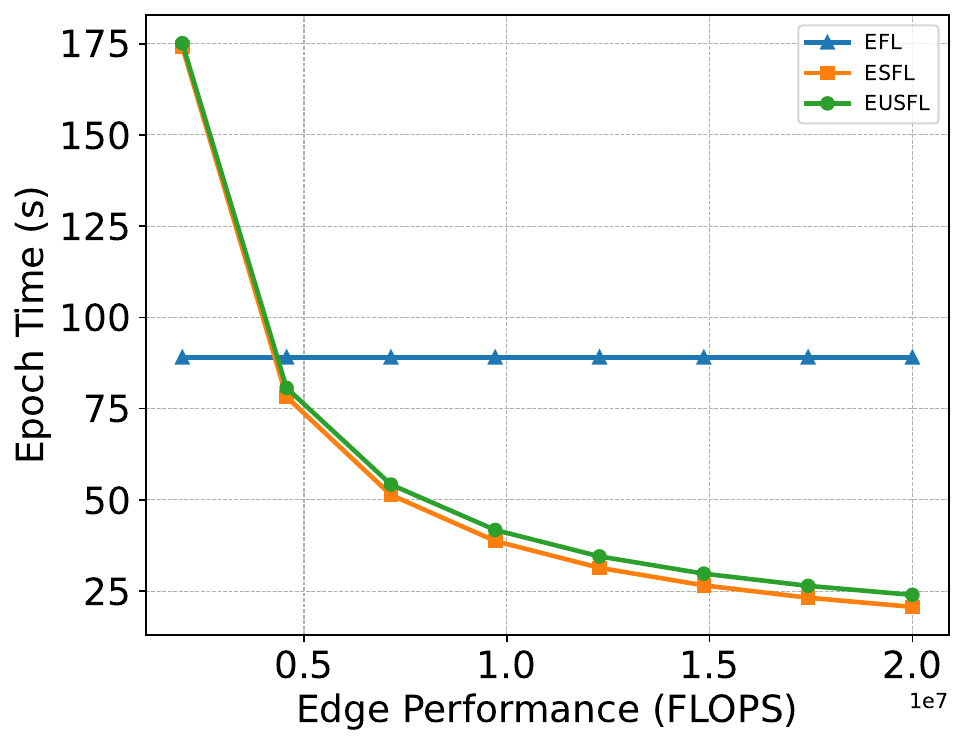}}
    \subfigure[]{\includegraphics[width=0.475\linewidth]{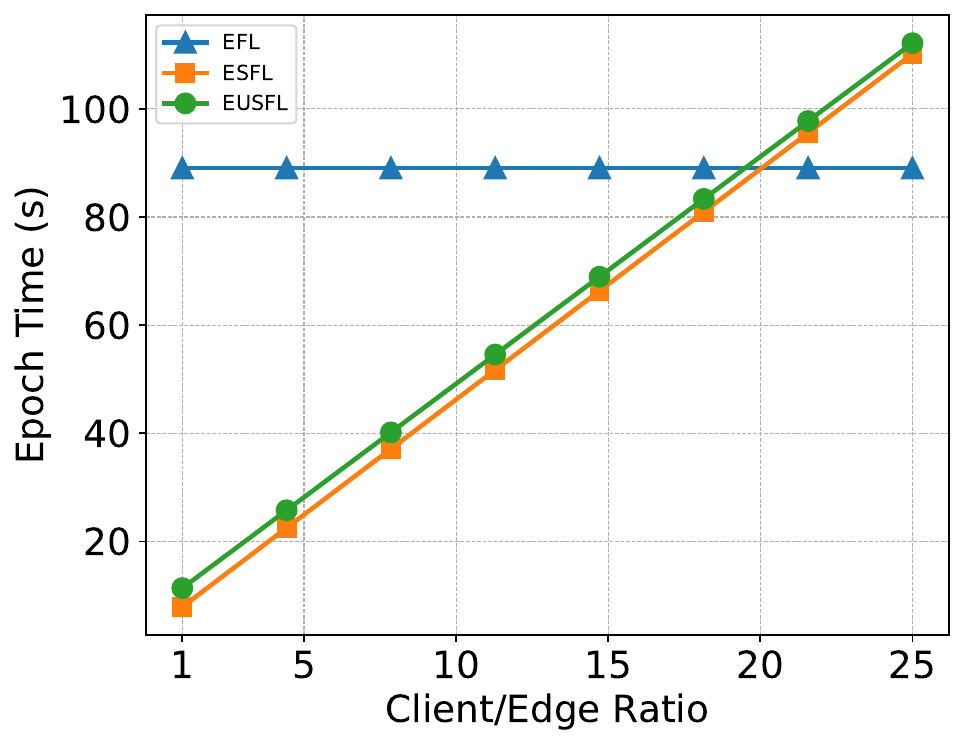}}
    \caption{Edge server Comparison: (a) Influence of Edge server Performance (b) Influence of Client/Edge server Ratio}
    \label{Fig_edge_comparison}
\end{figure}

Fig. \ref{Fig_edge_comparison} shows the influence of Edge server performance on single epoch training time, with all other factors held constant. As can be seen from Fig\ref{Fig_edge_comparison}(a), the training time of EFL is not affected by Edge server's performance, because Edge server does not participate in model training in EFL. Controversially, SFL schemes are greatly affected by Edge server performance. Because split model in at Edge server side, will correspond one-to-one with Client, the computing performance will be averaged. When Edge server's performance is high enough, the power allocated to each Client is also sufficient. To compare the influence of the number of clients subordinate to Edge server on epoch time, Fig\ref{Fig_edge_comparison}(b) shows how it changes with respect to the ratio of each Edge server to Client.

From the above two comparisons, it can be seen that there are two main factors affecting the epoch time, one is the computing performance of Client, and the other is the Client's transfer rate. Since the above two sets of comparative experiments only have one independent variable, it is not easy to obtain results for complex scenarios. Therefore, Fig. \ref{Fig_client_3d} is based on three main methods to compare the two factors. There are two independent variables in Fig. \ref{Fig_client_3d}, Client's computing performance and Client's transfer rate to Edge server. As shown from result, the training time reduced as computing performance and transfer rate increase. However, the degree of influence varies for different methods.

In Fig. \ref{Fig_client_3d}(a), FL is most affected by the Client's computing performance among the three methods because FL performs the whole training process locally. Therefore, when Client's computing capability is not enough, the process of training will take a long time. In Fig. \ref{Fig_client_3d}(c), U-shaped SL is most affected by Client's transfer rate because of the high communication requirement of U-shaped SL. In Fig. \ref{Fig_client_3d}(b), vertical SL performs best in the same scenario, but data privacy is not ensured. 

It can be seen that when computing capability is not enough but communication capability is sufficient, and strong privacy protection is required, using EUSFL to efficiently train the model is a very good choice

\subsubsection{Practical Scenario Simulation}
In Setup 1, we assume the following simulation scenario: each IoT device is connected to its Edge server through the network that can upload and download data. Considering privacy issues, IoT devices will not upload any raw data to the Edge server (including labels). The Edge servers are managed by the Central server, which is also connected to the Central that can also upload and download.

As Table \ref{table_Basic_Comparison} shows, SFL has the lowest time spent, ESFL is second, while EUSFL has slightly worse time spent than these two methods, but without the risk of leaking label information. Similarly, the time cost of EUSFL is much less than that of the centralized DL, FL, and EFL methods, which indicates that the related methods of SL can effectively improve the overall efficiency. Specifically, the edge-assisted EUSFL method, like other methods using SL, substantially reduces the amount of floating-point operations on the client side by splitting the training process of the neural network, thus achieving a reduction in time, while the Edge server with stronger performance effectively assists the overall training process.

\begin{table*}[tb]
\footnotesize
\centering
\caption{Setup 1 Result and Comparison}
\tabcolsep 4.2pt 
    \begin{tabular*}{0.99\linewidth}{lccccccccccccc}
    \toprule
      \multirow{2}{*}{Method} &
      \multicolumn{2}{c}{Client} &
      \multicolumn{2}{c}{Edge Server} &
      \multicolumn{2}{c}{Central server} &
      \multicolumn{2}{c}{C-E transfer} &
      \multicolumn{2}{c}{E-S transfer} &
      \multicolumn{2}{c}{C-S transfer} &
      \multirow{2}{*}{Total Time} \\
    \cmidrule(l{2mm}r{2mm}){2-3} \cmidrule(l{2mm}r{2mm}){4-5} \cmidrule(l{2mm}r{2mm}){6-7} \cmidrule(l{2mm}r{2mm}){8-9} \cmidrule(l{2mm}r{2mm}){10-11} \cmidrule(l{2mm}r{2mm}){12-13}      
         & FLOPS    & Time(s)  & FLOPS    & Time(s)   & FLOPS    & Time(s)   & Transfer & Time(s)    & Transfer & Time(s)    & Transfer & Time(s)     &          \\
    \midrule
    DL     & 0        & 0     & 0        & 0      & 35540000 & 2.96 & 0        & 0      & 0        & 0      & 54880000 & 2744.00    & 2746.96 \\
    FL     & 35540000 & 88.85 & 0        & 0      & 0        & 0     & 0        & 0      & 0        & 0      & 88192    & 4.41  & 930.26  \\
    SFL    & 1440000  & 3.60   & 0        & 0      & 34100000 & 2.84 & 0        & 0      & 0        & 0      & 14016    & 0.70  & 70.14   \\
    USFL  & 1940000  & 4.85  & 0        & 0      & 33600000 & 2.80   & 0        & 0      & 0        & 0      & 1437392  & 71.87 & 790.52  \\
    EFL    & 35540000 & 88.85 & 0        & 0      & 0        & 0     & 88192    & 0.22 & 88192    & 0.07 & 0        & 0       & 890.14  \\
    ESFL   & 1440000  & 3.60   & 34100000 & 4.26 & 34100000 & 2.80 & 14016    & 0.03 & 88192    & 0.07 & 0        & 0       & 100.81  \\
    EUSFL & 1940000  & 4.85  & 33600000 & 4.20    & 33600000 & 2.80   & 1437392  & 3.52 & 88192    & 0.07 & 0        & 0       & 150.45  \\ 
    \bottomrule
    \end{tabular*}
\label{table_Basic_Comparison}
\end{table*}

In Setup 2, we assume a scenario where several smart security cameras monitor multiple areas of an industrial area, and the cameras will continuously collect image information during the monitoring process. To construct a smart industrial area, it is necessary to train a model for identifying which type of object the image information is. However, due to the limitation of privacy, these images will not be transmitted to a certain center for centralized training but can only be trained locally using the camera. There are two training paradigms to consider at this point, EFL and EUSFL.

In this scenario, the camera uses an Arm Cortex-M55 processor and is equipped with a Linux operating system. Considering that the camera needs to maintain its basic functions, such as device control, access management, data transmission, etc., the computing performance that the camera can use for additional DL training is about 1 GFLOPS. That is, it can perform floating number operations $10^9$ times per second.

Each area is managed by an Edge server, and cameras in different areas are connected to different Edge servers through the WIFI protocol. When using WIFI for data transmission, the download and upload rates are about 10.12Mb/s and 6.12Mb/s (regarded as 8Mb/s transfer rate in the experiment). The Edge server is a high-performance computer with an i7-13700KF processor and an Nvidia RTX3080 GPU. Under this setting, the Edge server can independently and in parallel allocate 20G FLOPS of computing performance to each camera to assist it in DL model training.

With the mentioned experiment detail, the result is shown in Table \ref{Table_set2_2}. As known from Fig.\ref{Fig_models}, the computation of VGG-16 is very expensive, but over 99\% computational requirement could be split to the Edge server. In this condition, FL spent the longest time finishing the training task because of the heavy computational burden. Both the computation and communication requirements of Client are the least, so the time consumption of SFL is the shortest. Compared with SFL, USFL needs more data to be transferred due to a U-shaped split, and since the model of VGG-16 is very large, The time consumption also gets longer because ESFL needs an extra aggregation between the Edge server and the Central server. From the result, we can get to know, under such computation-limited and privacy-preserving conditions, using EUSFL can greatly shorten training time.

\begin{table*}[tb]
\footnotesize
\centering
\caption{Setup 2 Result and Comparison}
\tabcolsep 11pt 
    \begin{tabular*}{0.99\linewidth}{lccccc}
    \toprule
    VGG-CIFAR & Client Computation & Upper Entity Computation & Total Transfer (bytes) & Total Time (s) & Accuracy (10 epochs) \\ 
    \midrule
    FL     & 11,985,747,968 & 0              & 278,026,064 & 1402.116 & 64.80 \\
    SFL    & 130,023,424    & 11,855,724,544 & 541,680     & 12.813   & 65.12 \\
    USFL  & 132,644,864    & 11,853,103,104 & 18,177,408  & 151.873  & 64.92 \\
    EFL   & 11,985,747,968 & 0              & 278,026,064 & 409.468  & 65.03 \\
    ESFL   & 130,023,424    & 11,855,724,544 & 541,680     & 127.060  & 64.92 \\
    EUSFL & 132,644,864    & 11,853,103,104 & 18,177,408  & 128.881  & 65.13 \\
    \bottomrule
    \end{tabular*}
\label{Table_set2_2}
\end{table*}

\subsection{Discussion and Prospect}

According to the analysis and experimental evaluation of the EUSFL framework, it is shown that the framework has good scalability and performance for different FL methods, and effectively reduces the computational consumption of the client and the overall computational time spent while ensuring strong security and privacy. However, there are still some optimization directions for the EUSFL framework. The following directions can further enhance the training process of EUSFL.

\subsubsection{Model Design}

From the Efficiency Analysis chapter, it can be seen that one of the directions to optimize the training time is to optimize the smashed data transmitted. One of the directions to optimize the intermediate data is to further design the layers of the neural networks. For example, to design convolution and pooling layers in appropriate places, and the use of convolution cores to further reduce the smashed data that needs to be transmitted. If the size of smashed data is reduced, the communication overhead can be reduced, and the time required to complete the training can be shortened. Additionally, the smashed data output by different layers may not be the same, so designing the split mechanism $\mathcal{M}$ can also reduce communication overhead.

\subsubsection{Data Compression}

Compressing smashed data is another solution to reduce communication overhead. Although data compression may result in accuracy loss and decrease the convergence speed of the model, model compression can still be a viable option for improving the overall training speed if the decrease in communication overhead can be utilized to compensate for the number of training iterations needed to maintain overall accuracy.

\subsubsection{Asynchronous Training}

In FL, asynchronous training strategy can improve communication stability between devices and the Central server, as devices can update their models independently. This can reduce network congestion and latency, and enable FL in low-bandwidth or intermittent connectivity environments. Similarly, we can use asynchronous training to spread the computational burden of Edge to shorten the training process.

\section{Conclusion}

For the privacy and efficiency issues in training deep learning models in IoT environments, an efficient Edge-Assisted U-Split Federated Learning framework with privacy-preserving, EUSFL, is proposed to facilitate the efficient and collaborative training. EUSFL ingeniously apply unique U-Split method with the federated learning distributed architecture with additional LabelDP, our framework resisted data reconstruction attacks while ensuring data security. The U-shaped splitting method eliminates the need for any label information transmission during the training process, thus greatly reducing the threat of privacy leakage. LabelDP is also proposed to resist label reconstruction attack. The effectiveness of the EUSFL framework is verified through extensive experiments in simulation settings, and the performance is evaluated and analyzed in detail. When the communication conditions are favorable, EUSFL can effectively reduce the computational expenditure of devices, which in turn improves the efficiency of the whole system. Therefore, EUSFL is a promising solution in resource-constrained IoT environments, with good performance and scalability for different IoT scenarios.

\section*{Acknowledgments}
This work was supported by the Beijing Municipal Social Science Foundation (22GLC056).

\bibliographystyle{elsarticle-num.bst}
\bibliography{main}

\end{document}